\PassOptionsToPackage{table}{xcolor}
\documentclass[10pt,twocolumn,letterpaper]{article}

\usepackage[pagenumbers]{wacv} % To force page numbers, e.g. for an arXiv version

\usepackage{physics}
\usepackage{siunitx}
\usepackage{multirow}
\usepackage[dvipsnames,table]{xcolor}

\newcommand{\STAB}[1]{\begin{tabular}{@{}c@{}}#1\end{tabular}}

\definecolor{wacvblue}{rgb}{0.21,0.49,0.74}
\usepackage[pagebackref,breaklinks,colorlinks,allcolors=wacvblue]{hyperref}

\def\wacvPaperID{1147} % *** Enter the WACV Paper ID here
\def\confName{WACV}
\def\confYear{2026}

\title{Surf-NeRF: Surface Regularised Neural Radiance Fields}

\author{Jack Naylor \quad Viorela Ila \quad Donald G. Dansereau \\
Australian Centre for Robotics, School of Aerospace, Mechanical and Mechatronic Engineering, \\
The University of Sydney, NSW, 2006, Australia\\
{\tt\small {firstname.lastname}@sydney.edu.au}
}
\begin{document}
\maketitle
\begin{abstract}
Neural Radiance Fields (NeRFs) provide a high fidelity, continuous scene representation that can realistically represent complex behaviour of light. Despite works like Ref-NeRF improving geometry through physics-inspired models, the ability for a NeRF to overcome shape-radiance ambiguity and converge to a representation consistent with real geometry remains limited.  We demonstrate how both curriculum learning of a surface light field model and using a lattice-based hash encoding helps a NeRF converge towards a more geometrically accurate scene representation. We introduce four regularisation terms to impose geometric smoothness, consistency of normals, and a separation of Lambertian and specular appearance at geometry in the scene, conforming to physical models. Our approach yields 28\% more accurate normals than traditional grid-based  NeRF variants with reflection parameterisation. Our approach more accurately separates view-dependent appearance, conditioning a NeRF to have a geometric representation consistent with the captured scene. We demonstrate compatibility of our method with existing NeRF variants, as a key step in enabling radiance-based representations for geometry critical applications.\vspace{-1em}
\end{abstract}
    
\section{Introduction}
\label{sec:intro}
Neural Radiance Fields (NeRFs)~\cite{mildenhall2020nerf} provide an efficient coordinate based scene representation with wide applications to computer vision, robotics and beyond. The ray-based volumetric rendering formulation produces realistic novel views of a scene, encompassing view-dependent scene appearance. Complex appearances like specularity, transparency and interreflection lead to shape-radiance ambiguity where scene geometry is not uniquely represented, often producing imagined geometries.

\begin{figure}[htbp!]
    \centering
    \includegraphics[width=0.42\textwidth]{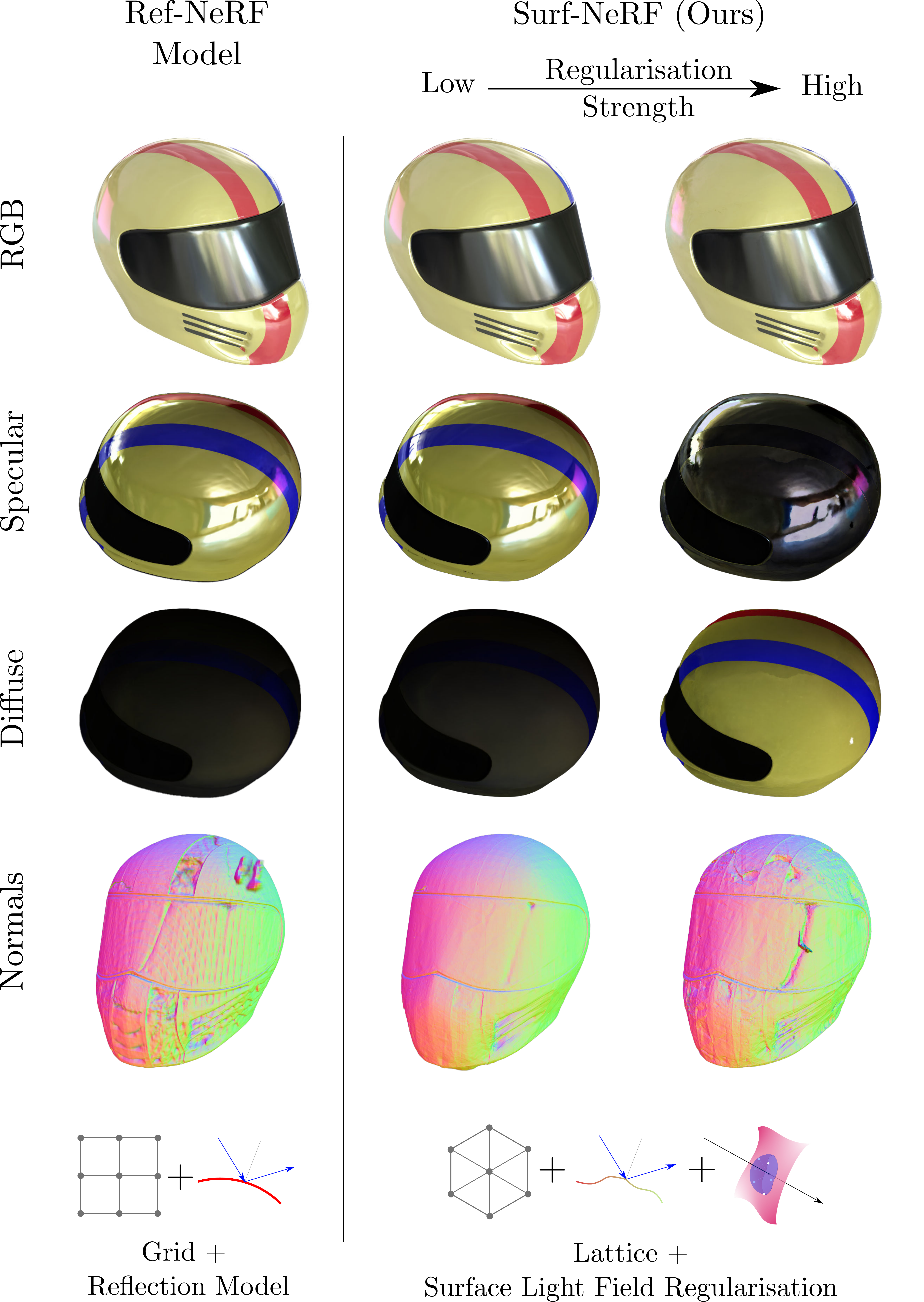}
    \caption{Surf-NeRF uses a permutohedral lattice encoding and the properties of surface light fields to regularise local regions of a NeRF. This improves consistency and smoothness in these regions by accumulating density at a surface, leading to improved geometry and more physically viable appearance components (ours, right) compared to current state-of-the-art (\cite{verbin2022ref}, left). Changing the frequency of regularisation balances between geometric fidelity and appearance separation.}\vspace{-\floatsep}
    \label{fig:fig1}
\end{figure}%
Re-parameterisation of the radiance in the scene as in Ref-NeRF~\cite{verbin2022ref} has shown improvements over previous state-of-the-art variants, however there exists an incomplete separation of the scene's Lambertian appearance from the specular and no explicit constraint on the placement of density. It is still common for the NeRF to place regions of density behind surfaces or in front of the camera as detached geometry (\textit{floaters}) to explain complex phenomena. This leads to a non-realistic geometric scene representation, as in Figure~\ref{fig:fig1}. 

Applications like robotics and 3D modelling require both accurate geometry and realistic appearance. 
Poor representation of physical geometry currently poses a large hurdle to the widespread adoption of radiance-based scene representation for geometry critical tasks. Robotic manipulation, autonomous navigation and mapping require an accurate scene structure to minimise the gap between a representation and the real world, for example when grasping metal objects or navigating near reflective windows. Similarly, traditional structure-from-motion pipelines have great difficulty in reconstructing visually complex objects including reflective and transparent surfaces.

We address geometric inaccuracy arising from view-dependent phenomena with the insight that there are multiple formulations of the plenoptic function which can produce viable scene representations. A surface light field~\cite{woodsurface2000} describes the plenoptic function as light originating from a geometrically smooth surface, which we encode using a permutohedral lattice hash encoding. Our insight lies in that we can push the NeRF towards a more geometrically accurate representation in the same rendering framework using additional geometric and appearance based regularisation via curriculum learning. We use a first surface assumption to describe the location of geometry in the scene. 

Using a second sampling of the NeRF at these points, we regularise density to produce smoothly-varying normals and thin, continuous sheets of density which more realistically represent scene geometry. Regularising the view-dependent appearance to remove Lambertian components we reduce shape-radiance ambiguity and encouraging more correct structure of the surface. In summary, we make the following contributions:
\begin{enumerate}
\item We devise a novel regularisation approach which uses the structure of a neural radiance field to sample density, normals and appearance in the vicinity of geometry in the scene, allowing for additional representation-driven regularisation terms to be applied.
\item We apply local regularisation consistent with a surface light field radiance model, including geometric smoothness of density, local consistency of normals and a physically correct separation of Lambertian and specular appearance using a light interaction model.
\item We introduce a permutohedral lattice encoding for NeRF which better represents non-planar and complex geometry over traditional cubic hash grids.
\item We introduce the \textit{Koala} dataset of four scenes with visually challenging objects and complex geometry to benchmark our approach.
\end{enumerate}

We benchmark our approach on state-of-the-art physics based NeRF variants, however our methodology may also be applied to other NeRF frameworks and likely adapted to other radiance based representations. 

This work is a key step in the deployment of NeRFs as a scene representation where both geometric and visual fidelity are critical, like robotic manipulation and navigation in complex unstructured environments.

\section{Related Work}

\textbf{Neural Scene Representations:} Implicit neural field approaches~\cite{irshad2024neuralfieldsroboticssurvey} produce continuous and high visual fidelity scene representations. These can be queried anywhere within the training set, balancing visual and geometric fidelity. In this work, we improve the geometry of a NeRF and maintain visual fidelity, reducing reliance on introduced geometry to meet scene appearance.

NeRFs~\cite{mildenhall2020nerf} leverage the efficiency of ray-based construction similar to light fields with a volumetric scene representation of points which emit light. Subsequent works improved the fidelity of representation~\cite{barron2021mipnerf,barron2022mip,Barron_2023_ICCV}, recovery of a static scene~\cite{sabour2023robustnerf, tancik2022block, martin2021nerf} and  performance around view-dependent phenomena such as reflections~\cite{verbin2022ref, guo2022nerfren}. However, relying on a purely volumetric rendering approach allows the network to imagine geometries to explain view-dependent phenomena, particularly in cases where regions of the scene are underconstrained~\cite{rebain2022lolnerf, kim2022infonerf, niemeyer2022regnerf} or where light does not follow the physical model of a straight ray through the scene~\cite{bemana2022eikonal}. Additional regularisation terms can dramatically improve the quality of rendering in these cases~\cite{niemeyer2022regnerf, kim2022infonerf}, e.g. improving the accuracy of normals significantly helps representation of complex appearances~\cite{ma2023specnerf, verbin2022ref}. 

More recent works~\cite{verbin2024nerf} have shown that ray-tracing multiple reflection rays can improve representation reflective scene element, albeit with extremely high computation cost. In this work we use properties of surface light fields to reduce the reliance of NeRFs on additional geometries, maintaining a single-pass volumetric rendering approach. %This approach explains complex visual phenomena entirely through view-dependent appearance, whilst maintaining the formulation of a single-pass volumetric rendering approach and its quality. We achieve this using a surface light field model and local characteristics of geometry and appearance of the scene.

\textbf{Radiance Based Scenes:} Recent advances in explicit representations like 3D Gaussian splatting~\cite{kerbl20233d} (3DGS) have demonstrated improved rendering quality over volumetric implicit models, reflective objects rely heavily on introduced Gaussians behind surfaces to represent view dependence. Although subsequent work has sought to improve the geometric fidelity of the resulting representation~\cite{huang20242d,jiang2025geometry,li2024geogaussian,wei2024normal,wang2024learning}, these involve substantial changes to the underlying representation or simplified geometry models due to the sparsity of the explicit radiance field. Our approach derives key benefits from the continuity of implicit radiance fields to formulate regularisation terms. We anticipate similarly inspired terms might be formulated for explicit methods like 3DGS, Radiant Foam and others~\cite{kerbl20233d,govindarajan2025radiant,liu2025deformable,huang20242d}.

\textbf{Surface Representations:} Signed distance fields (SDFs) have garnered significant attention~\cite{yariv2021volume, yu2022monosdf, dogaru2022sphere, fu2022geoneus, azinovic2022neural, oechsle2021unisurf} as they can represent smooth and continuous geometries in space and may be generated from multi-view constraints alone. %Applying view-dependent colour channels to the SDF~\cite{vicini2022differentiable, yariv2021volume, fu2022geoneus, wang2021neus, wang2022neus2} in a similar fashion to NeRFs has demonstrated improved accuracy and continuity of representation.
For diffuse planar objects, this representation provides a smooth and high-quality reconstruction; however, more highly view-dependent appearance and complex geometry (thin structures, concave geometries) are not well represented~\cite{wang2022neus2,rosu2023permutosdf,ge2023ref}. Alternate volumetric encodings better suited to complex geometry like PermutoSDF~\cite{rosu2023permutosdf} have substantially improved this representation, which we leverage in this work.
Some works have included a reflection parameterisation~\cite{ge2023ref}, however, struggle to reconstruct 3D scenes with fine detail, thin structures, and sharp changes compared to volumetric scenes. Appearance and geometry have an intrinsic relationship within neural representations, with geometry being altered to account for non-Lambertian appearance~\cite{fu2022geoneus}. We maintain a volumetric representation allowing for complex geometry, but with separated view-dependent and -independent appearance. %Given NeRFs are a high fidelity visual representation, we are concerned only with introducing a surface-like structure to the density field through volumetric rendering. This improves the geometry and consistency of novel view rendering by considering the NeRF as learning density constrained to a surface with smooth view-dependent terms.

\textbf{Appearance Vs. Geometry:}
Inverse rendering with mesh-based models produces a definite~\cite{garcesintrinsic2022} separation of the scene's appearance, geometry and environment, which are necessary to create realistic renderings of scenes under new conditions. Performing this without prior knowledge of the scene proves to be an immensely difficult task~\cite{zhang2021ners,boss2022samurai, cheng2022diffeomorphic, yao2022neilf}, however, utilising physically-based rendering fused with learnt appearance~\cite{boss2021nerd, boss2021neural} provides sufficient constraints to recover the components of the scene.%

Radiance field approaches seeking to learn an accurate scene representation under few view scenarios have used depth consistency~\cite{nagabushan2023siggraphasia}, additional depth supervision~\cite{Zhu_2023_CVPR} and regularisation using priors~\cite{niemeyer2022regnerf}. Incorporating an understanding of the physical interactions of light within a scene and its effect on appearance~\cite{zhang2022iron} under a solid surface assumption enables accurate geometry to be learnt alongside appearance. In the presence of specularities and other visual phenomena it is difficult to disentangle where appearance has been baked into geometry~\cite{yang2022neumesh}. 

Without the need for recovering a bidirectional reflectance distribution function (BRDF), light field approaches~\cite{garces2017intrinsic} provide a more generalised framework. This enables a clear separation between appearance and geometry, reducing the need to acquire environmental view-dependent effects. Our work leverages similar light-field characteristics to recover geometry more accurately within the volumetric framework of NeRF, adding a prior to how geometry and appearance should present in the scene.%
\section{Method}
\begin{figure*}[htbp]
\centering
    \centering
    \includegraphics[width=\textwidth]{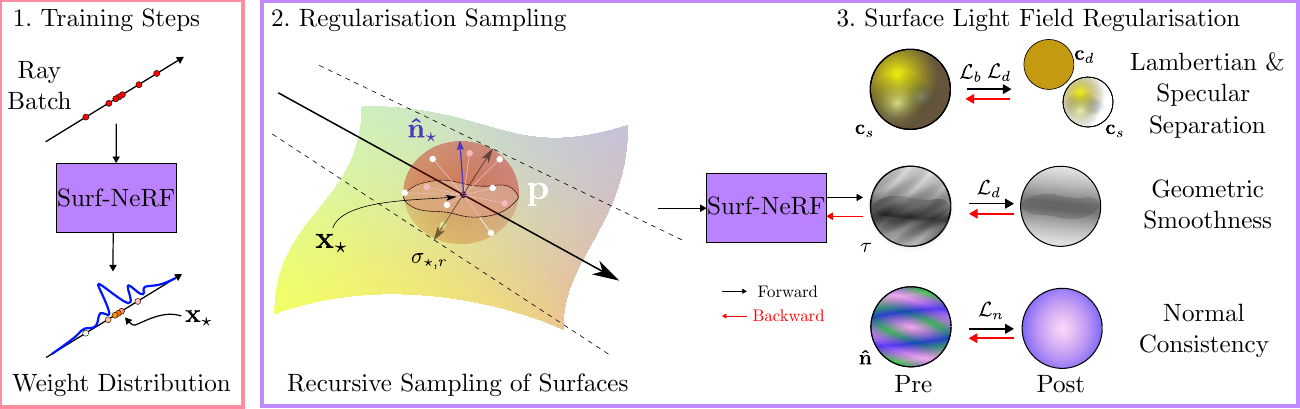}
    \caption{An overview of the Surf-NeRF methodology. We use a first surface assumption of light to locate samples $\vb{x}_\star$ which are likely to lie on a surface in the scene during regular training steps. Sampling at the surface in multiple directions, and at nearby points drawn from a sphere $\vb{p}$, we impose regularisation on directional and spatial behaviour in this region. We separate the Lambertian component from the specular colour channel $\vb{c}_s$ by using a total variation prior. Using neighbouring points we also regularise geometric smoothness on density $\tau$ and consistency of normals $\vu{n}$ leading to improved, continuous geometry.}
    \label{fig:sample_point_selection}\vspace{-\floatsep}%
\end{figure*}
Surf-NeRF introduces novel regularisation to locally enforce the properties of surface light fields. By regularising towards this representation, we represent smooth geometry with a physically viable separation of appearance and geometry producing more geometrically accurate radiance fields. This reduces shape radiance ambiguity by encouraging continuous regions of density with a smoothly varying view-dependent appearance. A visual depiction of our methodology is shown in Figure~\ref{fig:sample_point_selection}.
\subsection{Preliminaries}
A surface light field is a subset of the plenoptic function~\cite{bergen1991plenoptic} that provides the colour of a light ray originating from a surface within the scene. Solid scene geometries are well represented given the decoupled parameterisation of the scene geometry and radiance on these surfaces. 

A surface light field $L$ exists strictly on a surface geometry $G$ mapping directions in the unit 2-sphere ($\mathbf{S}^2$) to radiance, $L\,:\,G\times \mathbf{S}^2 \rightarrow \mathbf{c}$, for an RGB colour triplet $\mathbf{c}$~\cite{woodsurface2000}. 

Few real surfaces have an entirely view-dependent appearance. Similar to traditional light fields~\cite{garces2017intrinsic}, a surface light field may be decomposed into a Lambertian (diffuse) reflectance and a view-dependent component~\cite{woodsurface2000, Li_2018_ECCV}. %These intrinsic decompositions are defined for a point on the surface, $\vb{u}$, and viewing direction, $\pmb{\omega}$. The diffuse reflectance varies only with position over the surface, whilst the view-dependent component captures elements like reflection and refraction at the surface~\cite{garcesintrinsic2022}.
A surface light field has a strong assumption that the radiance seen from a given direction is provided entirely from a piecewise smooth geometry in the scene. This accordingly models phenomena such as volumetric scattering, reflection or transmission as a function of surface position rather than in a volumetric (NeRF) or physics-based (rendering engine) manner. In this way, a surface light field is decoupled from the geometry of a scene, meaning a surface may be deformed while its appearance is maintained~\cite{woodsurface2000}. Our approach is motivated by this decoupling to encourage Lambertian radiance to exist on smooth sheets of density, or surfaces, with a view-dependent colour.
\subsection{Model}
\textbf{Reflection Parameterisation:} Our method builds on the reflection parameterisation in Ref-NeRF~\cite{verbin2022ref}, which splits the scene into a diffuse and specular appearance term similar to a surface light field. This parameterisation provides enhanced results across Lambertian and specular scenes over original NeRF variants~\cite{mildenhall2020nerf,barron2021mipnerf} by learning a spatially varying diffuse colour $\vb{c}_d$, specular tint $\vb{s}$, rendering normal $\vu{n}'$ and a view-dependent specular colour $\vb{c}_s$ for each point in the scene. The final colour of a ray in this parameterisation is given as $\vb{c} = \vb{c}_d + \vb{s}\vb{c}_s$, a linear combination of the diffuse and specular terms. Ref-NeRF struggles to entirely separate Lambertian and specular appearance, utilising density instead to explain complex phenomena such as non-planar reflection, anisotropic reflection and interreflection, leading to the results seen in Figure~\ref{fig:fig1}. 

\textbf{Permutohedral Lattice Hash Encoding:} We replace the regular cubic hash grid used in prior NeRF variants~\cite{campzipnerf2024,mueller2022instant} with a hash-based permutohedral lattice~\cite{adams2010fast}. Prior work~\cite{rosu2023permutosdf}, has used this structure for its memory efficiency, but our insight lies in the volumetric coverage and relative density of simplices in this structure. A permutohedral lattice, is the most efficient lattice solution the sphere covering problem~\cite{conway2013sphere} in low dimensions. Vertices in this lattice are arranged evenly throughout $d$-space and packed more densely~\cite{adams2010fast}, allowing for more accurate interpolation of features over cubic volume elements. In $d=3$-space, simplices take the form of regular tetrahedra which can more accurately represent complex non-planar geometries and closely approximating the true geometry of the scene. Our permutohedral lattice encoding therefore has two main benefits over cubic grids: 1) half the memory accesses per interpolation by using barycentric interpolation of a tetrahedron~\cite{rosu2023permutosdf}, and 2) dramatically improved representation of non-planar surfaces derived from density by representing curved geometry with multi-resolution tetrahedral features.

We apply our regularisation losses denoted by $\mathcal{L}$ and permutohedral lattice encoding to ZipNeRF~\cite{Barron_2023_ICCV} using the Ref-NeRF parameterisation, leveraging its state of the art performance with grid-based encoding~\cite{mueller2022instant}. We make no modifications to the model itself beyond our hash encoding, but include a second, data-driven sampling to impose physically-inspired regularisation. We maintain two proposal networks and one NeRF network with 64 and 32 samples per ray. Importantly, our regularisation terms are extensible to other NeRF variants, and may be applied after the main training as a fine-tuning stage. An overview of model structure is provided in the supplementary materials.%, as we show in Section\TODO{ref{sec:results}}. 

\subsection{Sampling Radiance at a Surface}
\begin{figure}
    \centering
    \includegraphics[width=\linewidth]{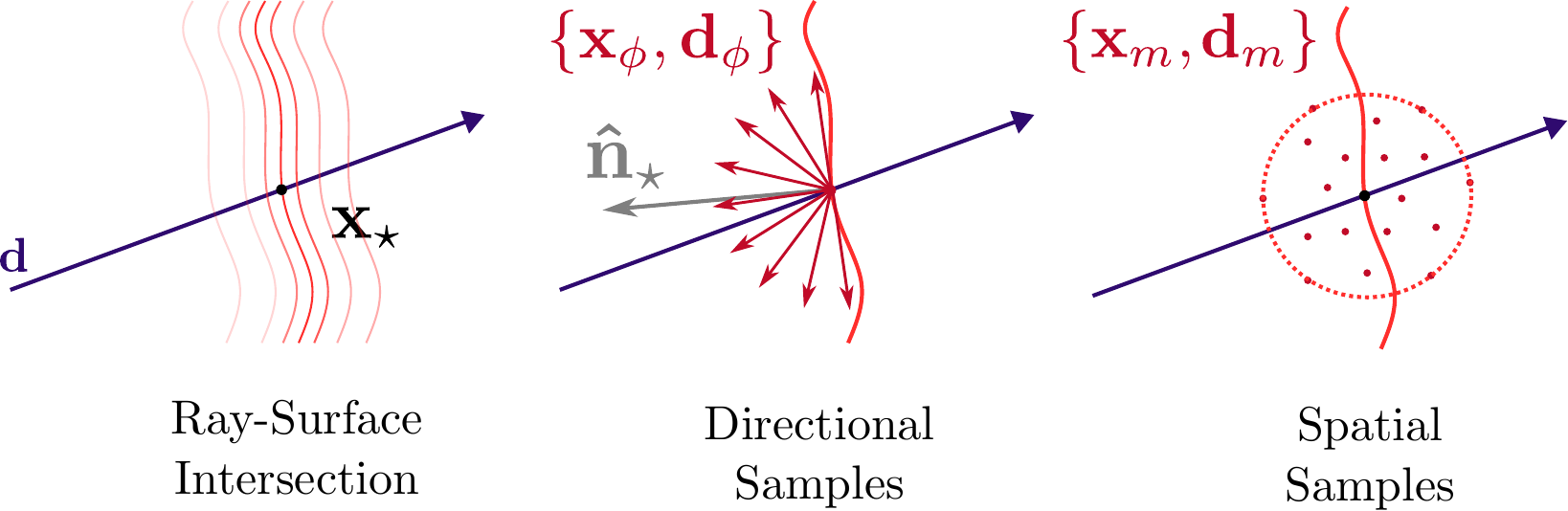}
    \caption{A visual depiction of the two sampling batches localised at a surface in the scene. After defining a ray surface intersection $\vb{x}_\star$, we use uniform samples on a unit ball to sample the surface through multiple viewing angles, and the local 3D volume.}
    \label{fig:sample_comparison}    \vspace{-\floatsep}%
\end{figure}

We regularise the surface and its light field by sampling a batch of positions and directions at the point where the ray intersects a surface in the scene, as shown in Figure~\ref{fig:sample_comparison}. These local regularisation terms are based on surface light field properties, current scene geometry and appearance at this location.

Where prior works have sampled unseen image patches~\cite{niemeyer2022regnerf} to encourage consistent depths, we sample a batch of unseen rays localised at a surface point in the scene to encourage local geometric continuity. This batch approach also has significant benefit over single perturbed points seen in prior work~\cite{oechsle2021unisurf, zhang2021nerfactor}, as it allows for changes in surface orientation and structure not captured by a single sample to be accounted for, as we detail in Section~\ref{sec:local_smooth}.

We utilise a first-surface assumption to infer the location of geometry; we assume the majority of radiance is emitted by the point of highest density closest to the camera. Our candidate surface is the first point along a ray with weight greater than the median weight of the ray, preventing regularisation from occurring behind the true location of the surface.  This minimises sampling around points which are occluded and therefore not well positioned with multi-view constraints. Choosing the median ensures that this selection is more robust to skewed weight distributions, particularly early during training. This is shown in Figure~\ref{fig:sample_point_selection} left.

Using the point $\vb{x}_\star$, ray origin $\vb{o}$ and direction $\vb{d}$, surface normal $\vb{\hat{n}}_\star$, and covariance $\vb{\Sigma}_\star$ of this sample's integrated positional encoding~\cite{barron2021mipnerf}, we generate two new batches: a spatial batch to regularise density and a directional batch to regularise appearance. Spatial samples query the density field $\tau$ in the local 3D region of the surface at $\vb{x}_\star$ and the directional samples query the distribution of view-dependent colour $\vb{c}_s$ at different viewing angles.

We adapt a deterministic sampling scheme on the sphere~\cite{frisch2023vmf_sampling} to produce uniformly distributed points used in both regularisation batches. Samples drawn from a von Mises-Fischer distribution with concentration parameter $\kappa=0$ are uniformly distributed on the unit 2-sphere $\mathbf{S}^2\subset \mathbb{R}^3$. We sample this distribution deterministically using a Fibonacci-Kronecker lattice as proposed by~\cite{frisch2023vmf_sampling} for $N$ samples. This is partitioned into $\log_2 N$ shells sampling the unit ball. Similar to the unscented sampling presented in ZipNeRF~\cite{Barron_2023_ICCV}, we apply a random rotation~\cite{avro1992fast} to these samples during training to avoid any bias from the orientation of the sampling scheme, arriving at samples $\vb{p}$. Further details are provided in the supplementary material.

We construct both batches of samples from virtual rays passing through this sphere centred at the surface. The points $p$ define directions $\vb{d}_s = \vu{p}$, origins $\vb{o}_s$ and covariances $\vb{\Sigma}_s$ to construct new conical frusta. The origins and covariances for these rays,
\begin{subequations}
    \begin{equation}
        \vb{o}_s = \vb{R}_s(\vb{o} - \vb{x}_\star) + \vb{x}_\star, \quad \vb{\Sigma}_s = \vb{R}_s\vb{\Sigma}_\star\vb{R}_s^\top, \tag{\theequation a,b}
    \end{equation}
\end{subequations}
are defined by the rotation matrix $\vb{R}_s$ which rotates the original ray direction $\vb{d}$ to each $\vb{d}_s$. This results in $N$ rays at the same distance to the surface as the original ray, with MipNeRF Gaussians~\cite{barron2022mip} aligned along ray directions.

Using these virtual rays, we construct sampling batches at positions $\vb{x}$, and directions $\vb{d}$ to employ surface light field regularisation. Both batches inherit covariances from the virtual rays $\vb{\Sigma}_s$ for anti-aliasing as in ZipNeRF~\cite{campzipnerf2024}.

\textbf{Directional Sampling:} Specularities at a surface are piecewise smooth, taking on the characteristics of what is being reflected. By sampling a point on a surface in a range of viewing directions and characterising the colour distribution, we can quantify how closely it matches the surface light field model introduced above. The directional batch $(\vb{x}_\phi, \vb{d}_\phi)$ samples at a single spatial location, but through a range of outward viewing angles:
\begin{equation}
    \qty(\vb{x}_\phi,\, \vb{d}_\phi) = \qty(\vb{x}_\star,\, \mathrm{sgn}(\vb{d}_s\cdot\vb{n}_\star)\vb{d}_s),
\end{equation}
where $\mathrm{sgn}(\cdot)$ is the sign function.

\textbf{Spatial Sampling:} In a NeRF, geometry is characterised by the density field $\tau(\vb{x})$, whose gradient with respect to position $\vb{x}$ approximates surface normals $\vb{n}\approx-\nabla_{\vb{x}} \tau$. By sampling density and normals of neighbouring points, and enforcing consistency between these values we can provide additional geometric supervision towards a surface model. The spatial batch $(\vb{x}_m, \vb{d}_m)$ consists of points $\vb{p}$ in the unit ball located at the surface $\vb{x}_\star$,
\begin{equation}
    (\vb{x}_m,\, \vb{d}_m) = \qty(\vb{x}_\star + \sigma_{\star,r} \vb{p}\sqrt{2\ln{2}},\, \vb{d}_s),
\end{equation}
where $\sigma_{\star,r}$ is the radial variance of the sample as per mip-NeRF~\cite{barron2021mipnerf}. This samples the NeRF within a single pixel's conical frustum, ensuring regularisation occurs at the scale of the training data; images closer to the scene regularise at finer detail. Our sample volume adapts in scale towards a minimum as the NeRF localises density during training during the later stages of training, as shown in the supplement. Since the colours go unused, the spatial batch uses the directions $\vb{d}_s$ of the virtual rays.

\subsection{Local Smoothness}~\label{sec:local_smooth}
We regularise the geometry of the representation by penalising sparse and irregular density, allowing for a smooth and continuous sheet-like density field to form. These constraints are valid except where topological edges or corners exist; we therefore formulate $L_1$-norm regularisation terms to allow for these local features to form when required. 

The spatial sampling batch encompasses the region in front of and behind the candidate surface. Using the normal at the candidate surface, we penalise points proportional to their density $\tau_j$ and perpendicular distance from the surface encouraging a plane of density to form perpendicular to the normal. Points far from the surface should have low density, and those on the same plane as the sample point should be dense. This loss term,
\begin{equation}
    \mathcal{L}_d = \lambda_d\sum_{\vb{r}\in\mathbf{R}}w_\star\sum_{j=0}^{N-1} \qty(1-e^{-\tau_j})\qty|\frac{(\vb{x}_{m,j}-\vb{{x}}_{\star})}{\norm{(\vb{x}_{m,j}-\vb{{x}}_{\star})}}\cdot\hat{\vb{n}}_\star|,
\end{equation}
sums over $N$ samples in the spatial batch and is also weighted by the surface point $w_\star$ rendering weight. This ensures that early during training, the NeRF is able to change surface locations as more rays sample the scene. We do not place a stop gradient on $\vu{n}_\star$, allowing for the surface to reorient based on the density of samples in the local volume.  

Normals in this local region should exhibit similar smoothness, and so the term,
\begin{equation}
    \mathcal{L}_n = \lambda_n\sum_{\vb{r}\in\mathbf{R}}w_\star\sum_{j=0}^{N-1}\qty(1-e^{-\tau_j})\frac{\qty(1-\qty(\hat{\vb{n}}_j\cdot\hat{\vb{n}}_\star))}{2},
\end{equation}
is imposed proportional to the angle between the normal at $\vb{x}_\star$ and those at each sample point $\vb{p}_j$. This term encourages dense samples to have a normal which is parallel with that at the surface point. Both terms localise density and encourage smooth, continuous geometries. We weight these terms by hyperparameters $\lambda_d$ and $\lambda_n$ respectively.
\subsection{Visually Realistic Appearance Separation}~\label{sec:specular}
Although highly specular surfaces or highly Lambertian surfaces are well represented with a reflection encoding, surfaces with both a strong Lambertian and specular component exhibit shape-radiance ambiguity. This is characterised by incorrect geometry with a view-dependent colour bias; that is the minimum specular colour over all viewing angles is non-zero to satisfy the appearance in all training view-points. More details are found in the supplementary.

To encourage the surface light field separation of the Lambertian and view-dependent terms we propose two additional losses based on directionally sampling a point. To encourage a minimum specular bias, we penalise the normalised specular colour over viewing angles in the directional sampling batch $\vb{c}_{s,\phi}$,
\begin{equation}
    \mathcal{L}_b = \lambda_b \sum_{\vb{r}\in\mathbf{R}} w_\star \sum_{j=0}^{N-1} \norm{\frac{\vb{c}_{s,j}}{\mathrm{sg}\qty(\norm{\max_\phi\,\vb{c}_s})}}^2,
\end{equation}
where $\mathrm{sg}(\cdot)$ is the stop gradient operator.
Where the specular distribution is uniform over viewing angles, this loss term is maximal. This ensures that the appearance of a point must accumulate most of its signal in either the specular or the Lambertian term - not in both. To maintain a piecewise smooth distribution in the specular, we enforce a total variation loss over the specular colour:
\begin{equation}
    \mathcal{L}_s = \lambda_s \sum_{\vb{r}\in\mathbf{R}} w_\star \sum_{j=0}^{N-1} \mathrm{STV}_j(\vb{d}_{\phi}, \vb{c}_{s}).
\end{equation}
This total variation is applied over the surface of the sampling sphere as a form of graph total variation. This term,
\begin{equation}
    \mathrm{STV}_j(\vb{d}_\phi, \vb{c}_{s}) = \sum_k \frac{1}{2} (\vb{d}_j \cdot \vb{d}_k+1)\norm{\vb{c}_{j} - \vb{c}_{k}}_1,
\end{equation}
is based upon the $k$-Nearest Neighbours of each sample point and weighted by the cosine distance between the sample and its neighbours. This approach provides a localised consistency which is edge-aware continuous over the sampling sphere, capturing occlusions. Both of these terms smooth the specular distribution and increase the reliance on the diffuse component of the model. These are weighted by hyperparameters $\lambda_b$ and $\lambda_s$ respectively.

\subsection{Curriculum Learning}
We impose our surface regularisation under curriculum learning to maintain visual fidelity and improve geometry. As training progresses and additional regularisation is imposed more frequently, the representation is refined to satisfy the geometric and appearance characteristics of a surface light field.
The surface constraints presented in Sections~{\ref{sec:local_smooth}-\ref{sec:specular}} are of similar computational complexity to a regular training step, requiring a second pass through both MLPs and therefore taking close to double the time. By using curriculum learning we trade off the improvement to the representation with training time. 

We schedule regularisation to occur in a staircase schedule of powers of 2: every 512 iterations early in the training, down to every 4 iterations in the final stages. On this schedule, there is approximately a 25\% increase to training time.

For each regularisation sample, we add on each regularisation loss $\mathcal{L}$ to the losses for the combined ZipNeRF and Ref-NeRF model structure. More model implementation details are provided in the supplementary material, along with hyperparameter weights. 
\section{Results}\label{sec:results}
Our regularisation is implemented in JAX~\cite{jax2018github} and based on the ZipNeRF codebase~\cite{campzipnerf2024}. We utilise the improvements to quality and speed in ZipNeRF combined with the physics-based model structure introduced in Ref-NeRF. We compare to prior works MipNeRF360~\cite{barron2022mip} and Ref-NeRF~\cite{verbin2022ref}.

We evaluate Surf-NeRF on the Shiny Real and the Shiny Objects~\cite{verbin2022ref} datasets used in Ref-NeRF as the main benchmark for our approach. We also introduce a captured dataset consisting of 4 complex reflective objects under controlled illumination, called the Koala dataset. Each scene contains approximately 40 training images and 10 test images with ground truth poses obtained using a Universal Robots UR5e robotic arm. We select objects that include non-planar specularities, fine details and specularities which present as virtual images in front of the surface. Further details are provided in the supplementary material.

\subsection{Shiny Objects Dataset}
Table~\ref{tab:blender_geom} summarises results on the Shiny Objects dataset. We compare visual fidelity by peak-signal-to-noise ratio (PSNR) in decibels (dB) and structural similarity scores (SSIM), and geometric fidelity by mean angular error (MAE) of surface normals in degrees and root mean squared error (RMSE) of the disparity in inverse units. We compare across positionally encoded methods (PE) based on MipNeRF360~\cite{barron2022mip}  and ``grid-based'' methods (Grid) based on ZipNeRF~\cite{Barron_2023_ICCV}. This includes baselines: MipNeRF with a diffuse channel (MipNeRF+Diff), and lattice and cubic grid based ZipNeRF with reflection parameterisation (Zip+Ref-NeRF). We train the PE implementations using V2-8 TPUs, and the Grid methods on four NVIDIA V100 32GB GPUs. Scene results are included in the supplementary materials.

%%%% Shiny objects results table
\begin{table}[htbp]
\centering
\caption{Summary on the \textit{Shiny Objects}~\cite{verbin2022ref} dataset between visual metrics (PSNR, SSIM) and geometric metrics (MAE, RMSE). Our method yields significantly more accurate normals, trading off a marginal decrease to PSNR. Yellow, orange and red are the third, second, and first scores.}~\label{tab:blender_geom}
\scriptsize
\begin{tabular}{clrrrr}
\hline
\multicolumn{1}{l}{} & Model            & PSNR $\uparrow$ & SSIM $\uparrow$ & MAE $\downarrow$ & RMSE $\downarrow$\\

                            \hline
\multirow{4}{*}{\STAB{\rotatebox[origin=c]{90}{PE}}}    & MipNeRF          & 31.29     &  0.943     &  56.17                         & 0.125                     \\
                            & MipNeRF+Diff     & 30.84     & 0.936     & 60.00                         &  \cellcolor{yellow!50}0.122                     \\
                            & Ref-NeRF         &   \cellcolor{red!50}33.21     &  \cellcolor{red!50}0.971     &  25.89                         &  \cellcolor{red!50}0.117                     \\
                            & Surf-NeRF (Reg. Only) &   33.01     &  \cellcolor{orange!50}0.967     &  22.15                         &  \cellcolor{red!50}0.117                     \\ \hline
                        \multirow{5}{*}{\STAB{\rotatebox[origin=c]{90}{Grid}}}    & ZipNeRF          &  31.02     &  0.952     &  19.47                         &  0.209                     \\
                            & Zip+Ref-NeRF      &  \cellcolor{yellow!50}33.14     &   \cellcolor{yellow!50}0.964     & 14.70                         &  0.195                    \\
                            & Lattice+Ref-NeRF      &  \cellcolor{red!50}33.21    &   \cellcolor{yellow!50}0.964     & \cellcolor{yellow!50}13.76                         &  0.202                    \\
                            & Surf-NeRF (No Lattice) &  32.13    &  0.954     &  \cellcolor{orange!50} 13.35                         & 0.202 \\
                            & Surf-NeRF (Ours) &  32.82    &  0.955     &   \cellcolor{red!50}10.60                       & 0.198\\
                            \hline
\end{tabular}   \vspace{-0.5\floatsep}%

\end{table}

Across both the positional encoding based and hash based networks we see comparable performance to Ref-NeRF in visual fidelity, demonstrating our ability to maintain visual fidelity and encourage a more geometrically accurate representation. In the positionally encoded approaches we see a 14.4\% improvement to normal accuracy, with notable improvements around regions of high specularity. Surf-NeRF attains a 27.9\% improvement to normals for the grid based implementation. Given the discretisation of the grid-based approaches, we see comparatively higher disparity errors across the dataset but improved normals from the interpolation of encodings.

\begin{figure}[htbp!]
    \centering
    \includegraphics[width=\linewidth]{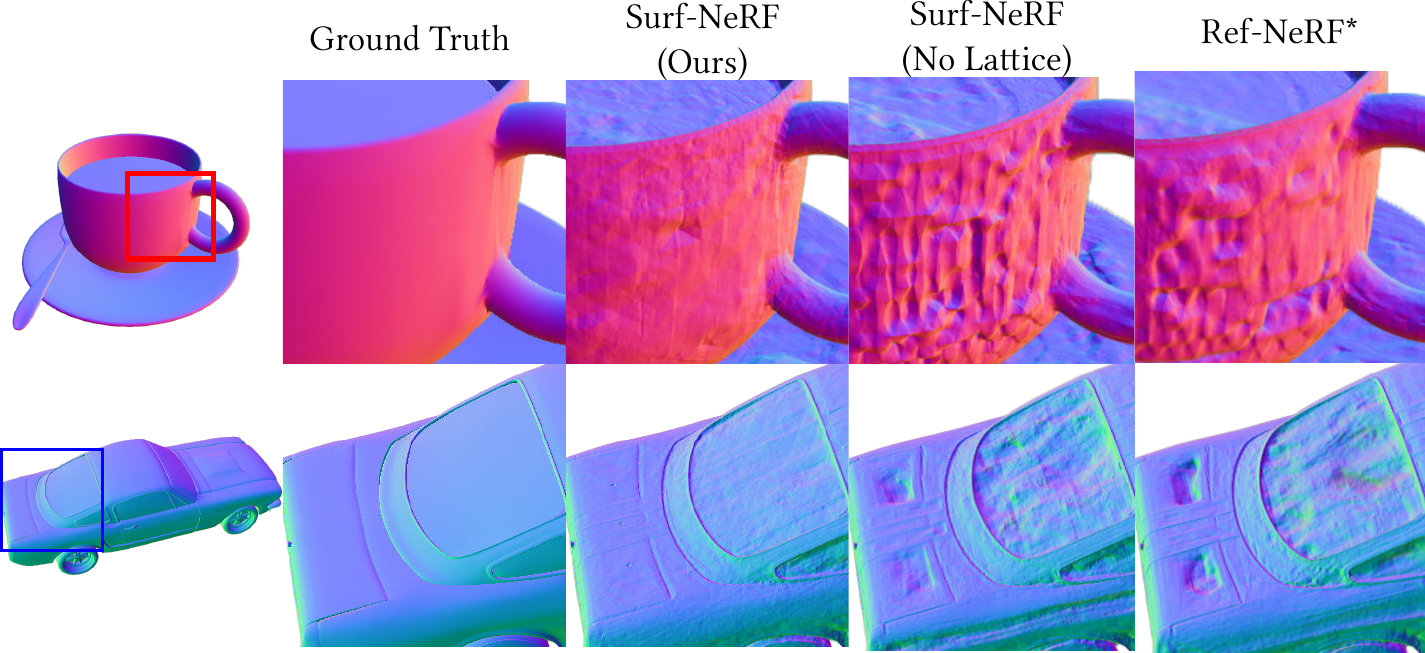}
    \caption{Geometry comparison between normals achieved by Surf-NeRF and traditional grid implementations on the car and coffee scenes. Ref-NeRF* indicates the Zip+Ref-NeRF model. Regularising the density field improves the continuity of surfaces reducing holes and divots. The inclusion of the permutohedral lattice encoding significantly improves surface normals by better approximating smooth surface curvature.}
    \label{fig:geometry_improvements}\vspace{-0.5\floatsep}
\end{figure}

\begin{figure}[htbp!]
    \centering
    \includegraphics[width=\linewidth]{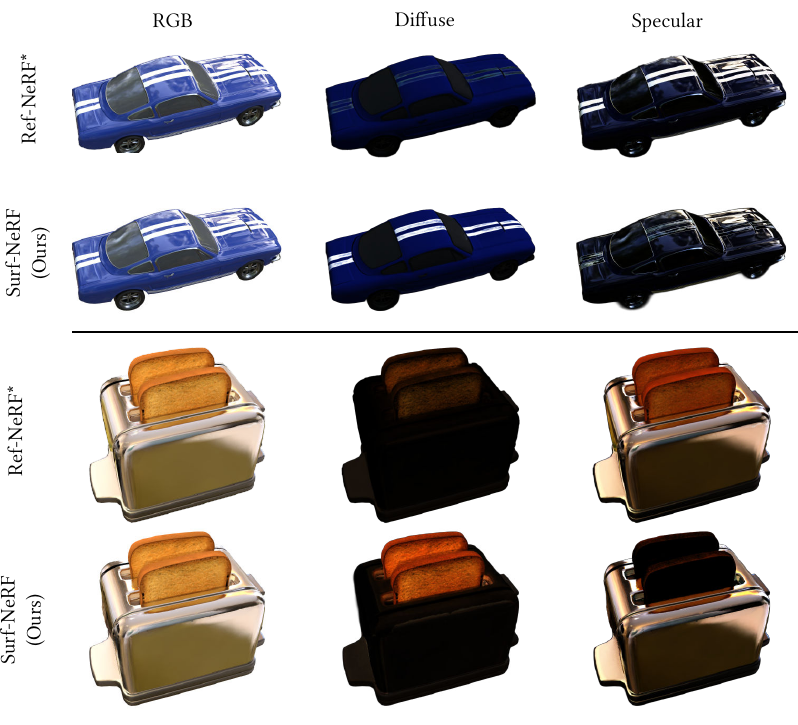}
    \caption{Appearance separation between  Surf-NeRF and baseline method Zip+Ref-NeRF. Colour renderings are comparable, but Surf-NeRF produces physically based separation of Lambertian and specular scene content, such the racing stripes and toast.}
    \label{fig:appearance_improvements}\vspace{-0.5\floatsep}
\end{figure}

In Figure~\ref{fig:geometry_improvements} we demonstrate that the proposed regularisation is able to substantially improve surface normals and the surface consistency. This also separates the Lambertian scene content, including the racing stripes as in \ref{fig:appearance_improvements}. Similarly, on the toaster scene we successfully separate the Lambertian toast and retain its reflection in the specular term.

% \begin{figure*}[htbp]
%     \centering
%     \includegraphics[width=\textwidth]{images/refnerf_subsets.pdf}
%     \caption{Results from the grid based models. By regularising the scene, we not only remove Lambertian scene content from the specular, like the stripes on the car and toast in the toaster, but our approach yields improved geometry particularly around curves and specularity, like the car windshield and corners of the toaster. Ref-NeRF* indicates the Zip+Ref-NeRF model.}
%     \label{fig:grid_based_improvements}\vspace{-\floatsep}
% \end{figure*}

% Figure~\ref{fig:teapot} depicts performance of positionally encoded models on specular surfaces with low texture. Surf-NeRF yields drastically better surface geometry and normals.

% \begin{figure}[htbp]
%     \centering
%     \includegraphics[width=\linewidth]{images/teapot_pe.pdf}
%     \caption{Geometry results on the non-grid based models. The surface regularisation can resolve errors in geometry which stem from shape-radiance ambiguity e.g. in regions with low-frequency textures on the teapot lid, without compromising visual fidelity. This attains substantially improved normals in these regions.}
%     \label{fig:teapot}\vspace{-\floatsep}
% \end{figure}

\subsection{Shiny Real Dataset}
The visual fidelity results on the \textit{Shiny Real} dataset~\cite{verbin2022ref} are shown in Table~\ref{tab:shiny_real} for Surf-NeRF and comparison baseline models. We demonstrate comparable visual fidelity to other state of the art NeRF variants demonstrating no appreciable drop in visual fidelity compared to non-regularised approaches on the positional encoding based models. The complex appearances of these datasets combined with the hash encoding sees a small drop in PSNR values for grid-based variants. Surf-NeRF shows decreased reliance on floaters to explain appearance as shown in the supplement.

\begin{table}[htbp]
\centering
\caption{Summary of numerical results for captured \textit{Shiny Real} dataset~\cite{verbin2022ref}. Surf-NeRF maintains visual fidelity on captured scenes, improving SSIM over the PE based methods.}~\label{tab:shiny_real}
\scriptsize
\begin{tabular}{clrr}
\hline
& Model                 & PSNR           & SSIM          \\
\hline
\multirow{4}{*}{\STAB{\rotatebox[origin=c]{90}{PE}}}  &MipNeRF          & \cellcolor{red!50}23.72          &0.543         \\
&MipNeRF+Diff     & 23.62          & 0.543         \\
&Ref-NeRF         & \cellcolor{orange!50}23.65          & 0.529         \\
&Surf-NeRF (Reg. Only) & 23.26          & 0.573         \\\hline
\multirow{5}{*}{\STAB{\rotatebox[origin=c]{90}{Grid}}}  &ZipNeRF          & \cellcolor{yellow!50}23.63          & \cellcolor{red!50}0.626         \\
&Zip+RefNeRF      & 22.63          & \cellcolor{red!50}0.626         \\
&Lattice+RefNeRF      & 22.94          & \cellcolor{orange!50}0.614         \\
&Surf-NeRF (No Lattice) & 22.48        & 0.600         \\
&Surf-NeRF (Ours) & 23.04        & 0.600        \\\hline
\end{tabular}    \vspace{-\floatsep}%
\end{table}

\subsection{Koala Dataset}
\begin{figure}[htbp!]
    \centering
    \includegraphics[width=\linewidth]{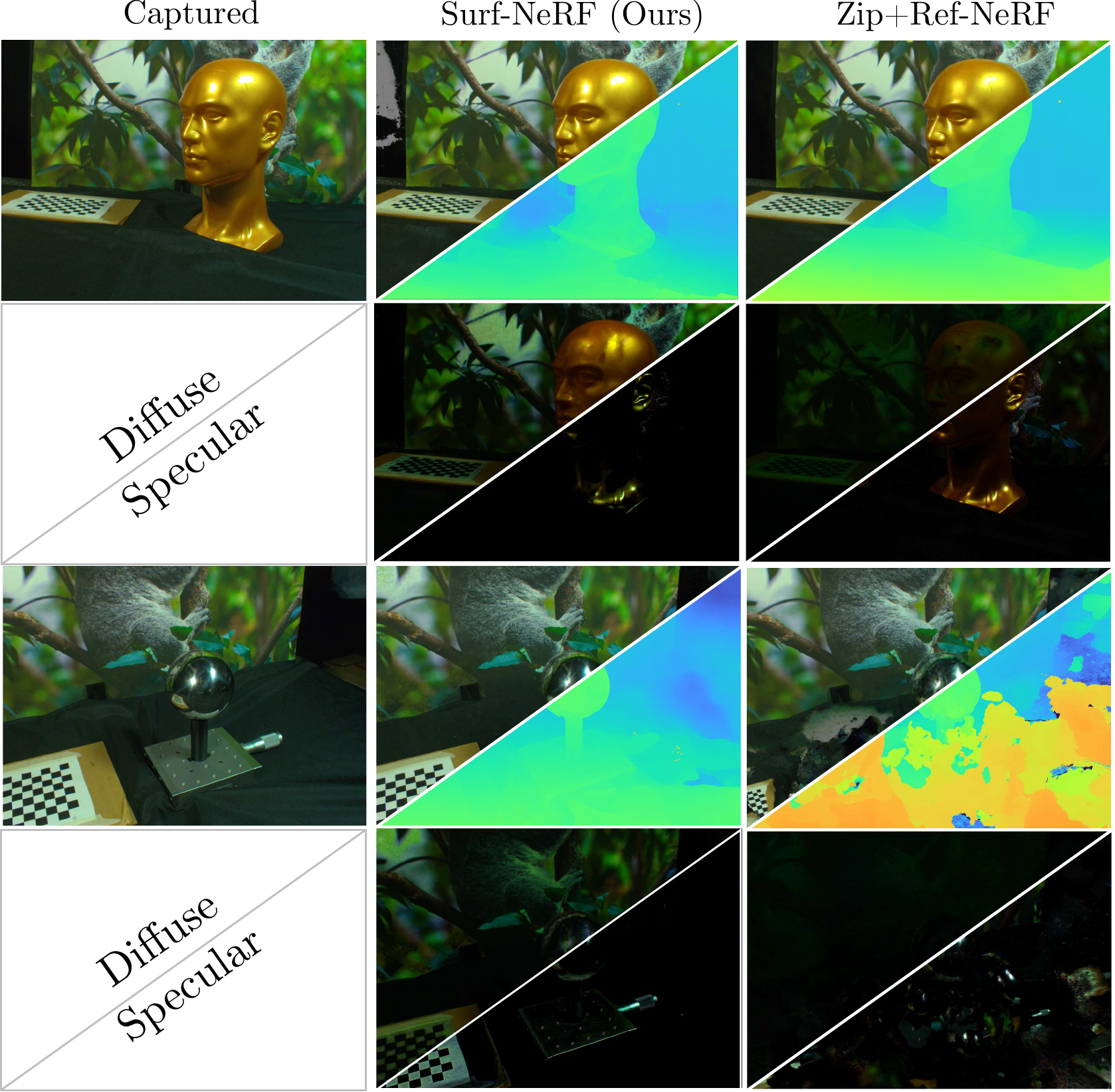}
    \caption{The ``Gold Head'' and ``Shiny Ball'' scenes from the \textit{Koala} dataset. Top rows contain RGB and median depth renderings. The Gold Head scene shows complete separation of Lambertian and specular appearance for the mannequin. The Shiny Ball dataset represents a failure case for Zip+Ref-NeRF baseline, however the proposed regularisation successfully reconstructs and appropriately places the majority of the colour into the diffuse channel.}
    \label{fig:koala}\vspace{-0.5\floatsep}%
\end{figure}

Table~\ref{tab:koala} shows the strength of the proposed regularisation on complex reflective scenes. Figure~\ref{fig:koala} demonstrates the ability for the proposed regularisation to separate Lambertian and specular colours in a more physically consistent manner, and to overcome failure cases where specularities are not densely sampled by the training data. Additional results are provided in the supplementary.

\begin{table}[htbp!]\vspace{-0.5\floatsep}
\centering
\caption{Summary of numerical results for captured \textit{Koala} dataset. Surface regularisation helps scene convergence around complex specular geometry improving PSNR.}~\label{tab:koala}
\scriptsize
\begin{tabular}{lrr}
\hline
 Model                & PSNR        & SSIM        \\
\hline
ZipNeRF          & 27.79       & 0.663       \\
\hline
Zip+RefNeRF      & 23.69       & 0.632       \\
Surf-NeRF (No Lattice) & \textbf{27.24}       & \textbf{0.655}\\
Surf-NeRF (Ours) & 26.86 & 0.654 \\\hline
\end{tabular}    \vspace{-\floatsep}%
\end{table}

\subsection{Ablation Study}
As shown in Table~\ref{tab:ablation} including our full approach provides strong performance across both the visual and geometric metrics on \textit{Shiny Objects}~\cite{verbin2022ref}. Our appearance based regularisation is coupled closely to the geometry of the scene. The permutohedral lattice encoding significantly improves surface normals coupled with our regularisation. %Removing the specular bias term drastically increases the normal error, however does not alter disparity in the scene. The normal term improves continuity of the geometry in the scene. Removing this allows for each individual normal to be determined more freely, reducing the MAE, but it also allows for floaters to be introduced unnecessarily by our spatial term corresponding to an increase in PSNR but also an increase in disparity error within the scene.

\begin{table}[htbp!]\vspace{-0.25\floatsep}\centering
\centering
\caption{Ablation study on Surf-NeRF regularisation terms on the \textit{Shiny Objects} dataset. All terms contribute to overall performance. $\bigstar$ ablates deterministic sphere sampling, \# ablates the permutohedral lattice and all $\mathcal{L}$ terms match those in the paper.}~\label{tab:ablation}
\scriptsize
\begin{tabular}{l@{}c@{}c@{}c@{}c@{}c@{}crrrr}
\hline
 & $\bigstar$& \# & $\mathcal{L}_s$ &$\mathcal{L}_n$& $\mathcal{L}_d$ & $\mathcal{L}_b$ & PSNR$\uparrow$ & SSIM$\uparrow$ & MAE$\downarrow$ & RMSE$\downarrow$ \\
 \hline
& \checkmark  & & $\,$ & \checkmark & \checkmark & \checkmark &  33.03& 0.962& 15.47&  \cellcolor{red!50}0.181\\
& \checkmark &  & \checkmark & $\,$ & \checkmark & \checkmark & 32.55& 0.959& 15.81& \cellcolor{yellow!50}0.192\\
& \checkmark & & \checkmark & \checkmark & $\,$ & \checkmark & 32.83&  0.961& 14.83& 0.200\\
& \checkmark & & \checkmark & \checkmark & \checkmark & $\,$ & \cellcolor{red!50}33.32&  \cellcolor{orange!50}0.965&  16.34&  0.196\\
& \checkmark  & \checkmark & $\,$ & \checkmark & \checkmark & \checkmark &  \cellcolor{yellow!50}33.12 & \cellcolor{yellow!50}0.964  & 12.21  &  \cellcolor{orange!50}0.183 \\
& \checkmark &  \checkmark & \checkmark & $\,$ & \checkmark & \checkmark &  32.76 & 0.958 &  11.40 & 0.196\\
& \checkmark & \checkmark & \checkmark & \checkmark & $\,$ & \checkmark & 32.96 &  0.964 & \cellcolor{yellow!50}10.90 &  0.199 \\
& \checkmark & \checkmark & \checkmark & \checkmark & \checkmark & $\,$ &  \cellcolor{orange!50}33.14 &   \cellcolor{red!50}0.969 &  \cellcolor{orange!50}10.74 &  0.201 \\
\hline
 Rand. Samples & &\checkmark & \checkmark & \checkmark & \checkmark & \checkmark & 32.26 & 0.954 & 13.70 & 0.203  \\
  (No Lattice) & \checkmark & & \checkmark & \checkmark & \checkmark & \checkmark & 32.13& 0.954& 13.35&   0.202\\
 (Ours) & \checkmark &\checkmark & \checkmark & \checkmark & \checkmark & \checkmark & 32.82& 0.955&  \cellcolor{red!50}10.60&   0.198\\
\hline
\hline
\end{tabular}    \vspace{-0.5\floatsep}%
\end{table}

\begin{table}\vspace{-0.25\floatsep}
\centering
\caption{Parameter study on regularisation frequency. There is a tradeoff to regularisation frequency between quality, time and effect, as in Figure~\ref{fig:fig1}. The 512/4 regime balanced multi-view photometric losses with the surface light field regularisation.}~\label{tab:currablation}
\scriptsize
\begin{tabular}{c c r r r r}
\hline
Initial Freq. & Final Freq. & PSNR $\uparrow$ & SSIM $\uparrow$ & MAE  $\downarrow$ & RMSE  $\downarrow$ \\
\hline
1024 & 8 &  32.63 &  \textbf{0.955} &  12.46 & 0.204 \\
512 & 4 &  \textbf{32.82} &   0.953 & \textbf{10.60} &   \textbf{0.198} \\
256 & 2 &  30.59 &   0.946 & 18.99 &   0.220 \\
128 & 1 &   29.77 &   0.926 & 28.37 &  0.218 \\
\hline
\end{tabular}    \vspace{-0.5\floatsep}%
\end{table}

Table~\ref{tab:currablation} demonstrates a parameter study on our curriculum learning approach. Regularising frequency forces the NeRF to learn better separation of Lambertian and view-dependent channels, as in Figure~\ref{fig:fig1} but reduces overall geometric quality by strongly manipulating density earlier in training. See the supplementary for more detail.

\subsection{Surface Regularisation as Finetuning}
We demonstrate surface regularisation as a fine-tuning step, after a NeRF has been trained on pre-trained Zip+Ref-NeRF and ZipNeRF model on the car scene from the \textit{Shiny Objects} dataset. Since ZipNeRF has no Lambertian colour term, we do not include $\mathcal{L}_b$, and apply our sphere total variation regularisation $\mathcal{L}_s$ to the view-dependent colour $\vb{c}$, demonstrating applicability to models without reflection parameterisation. The spatial sampling terms depend only on the geometry and can be applied to other density-based neural fields without modification.  We continue training at a fixed learning rate of $3.25\times10^{-4}$ and present our results in Table~\ref{tab:finetuning}.  Normals are improved over the base model and na\"ive approach where the model continues training without our regularisation. Finetuning with the reflection parameterisation, as in our Zip+RefNeRF model slightly increases disparity errors as the NeRF changes density placement during the additional training.

\begin{table}[htbp]\vspace{-0.5\floatsep}
\centering
\caption{Surf-NeRF as a finetuning step with applicability to other NeRF variants. Additional training refines surface normals.}~\label{tab:finetuning}
\scriptsize
\begin{tabular}{llrrp{0.6cm}r}
\hline
\multicolumn{1}{l}{Base}    & Extra          & PSNR & SSIM  & Median MAE & RMSE  \\
\hline
\multirow{3}{*}{ZipNeRF}    & -              & \textbf{27.34}     &  \textbf{0.932} & 24.21      & 0.220 \\
                            & 2.5k ZipNeRF    & 27.32     & 0.931 & 24.30      & 0.220 \\
                            & 2.5k SurfNeRF   & 27.33     & 0.931 & \textbf{23.57}      & \textbf{0.218} \\\hline
\multirow{3}{*}{ZipRefNeRF} & -              & \textbf{30.25}     & \textbf{0.957} & 18.44      & \textbf{0.246} \\
                            & 2.5k ZipRefNeRF & 30.14     & 0.956 & 18.59      & 0.251 \\
                            & 2.5k SurfNeRF   & 30.13     & 0.956 & \textbf{18.29}      & 0.248\\\hline
\end{tabular}    \vspace{-\floatsep}%
\end{table}%

\subsection{Limitations} Modelling the scene as surfaces, volumetric elements like hair or subsurface scattering may have their representation degraded with the proposed regularisation. Our methodology takes twice as long on each regularisation step compared to a vanilla training step owing to two passes through the network. We enforce geometric consistency during training at the cost of reducing photometric accuracy, leading to slightly reduced PSNR scores.
\section{Conclusion}
We have proposed Surf-NeRF, a novel regularisation approach with four physically derived regularisation terms and a lattice hash encoding, and shown these to improve geometry in a NeRF by curriculum learning towards a surface light field model. This ensures smoothly varying surface normals and geometry with more realistic Lambertian and view-dependent appearance. This enables a NeRF to adjust geometric structure without greatly sacrificing visual fidelity, improving accuracy of up to 27.8\% in normals and 6.7\% in disparity. This work is a key step towards better radiance-based representations for geometry critical tasks.
{
    \small
    \bibliographystyle{ieeenat_fullname}
    \bibliography{main}

\begin{thebibliography}{56}
\providecommand{\natexlab}[1]{#1}
\providecommand{\url}[1]{\texttt{#1}}
\expandafter\ifx\csname urlstyle\endcsname\relax
  \providecommand{\doi}[1]{doi: #1}\else
  \providecommand{\doi}{doi: \begingroup \urlstyle{rm}\Url}\fi

\bibitem[Adams et~al.(2010)Adams, Baek, and Davis]{adams2010fast}
Andrew Adams, Jongmin Baek, and Myers~Abraham Davis.
\newblock Fast high-dimensional filtering using the permutohedral lattice.
\newblock In \emph{Computer graphics forum}, pages 753--762. Wiley Online Library, 2010.

\bibitem[Arvo(1992)]{avro1992fast}
James Arvo.
\newblock Fast random rotation matrices.
\newblock In \emph{Graphics Gems {III} {(IBM} Version)}, pages 117--120. Academic Press, 1992.

\bibitem[Azinovi{\'c} et~al.(2022)Azinovi{\'c}, Martin-Brualla, Goldman, Nie{\ss}ner, and Thies]{azinovic2022neural}
Dejan Azinovi{\'c}, Ricardo Martin-Brualla, Dan~B Goldman, Matthias Nie{\ss}ner, and Justus Thies.
\newblock {Neural {RGB-D} Surface Reconstruction}.
\newblock In \emph{Proceedings of the IEEE/CVF Conference on Computer Vision and Pattern Recognition}, pages 6290--6301, 2022.

\bibitem[Barron et~al.(2021)Barron, Mildenhall, Tancik, Hedman, Martin-Brualla, and Srinivasan]{barron2021mipnerf}
Jonathan~T Barron, Ben Mildenhall, Matthew Tancik, Peter Hedman, Ricardo Martin-Brualla, and Pratul~P Srinivasan.
\newblock Mip-{NeRF}: A multiscale representation for anti-aliasing neural radiance fields.
\newblock In \emph{Proceedings of the IEEE/CVF International Conference on Computer Vision}, pages 5855--5864, 2021.

\bibitem[Barron et~al.(2022)Barron, Mildenhall, Verbin, Srinivasan, and Hedman]{barron2022mip}
Jonathan~T Barron, Ben Mildenhall, Dor Verbin, Pratul~P Srinivasan, and Peter Hedman.
\newblock {Mip-NeRF} 360: Unbounded anti-aliased neural radiance fields.
\newblock In \emph{Proceedings of the IEEE/CVF Conference on Computer Vision and Pattern Recognition}, pages 5470--5479, 2022.

\bibitem[Barron et~al.(2023)Barron, Mildenhall, Verbin, Srinivasan, and Hedman]{Barron_2023_ICCV}
Jonathan~T. Barron, Ben Mildenhall, Dor Verbin, Pratul~P. Srinivasan, and Peter Hedman.
\newblock {Zip-NeRF: Anti-Aliased Grid-Based Neural Radiance Fields}.
\newblock In \emph{Proceedings of the IEEE/CVF International Conference on Computer Vision (ICCV)}, pages 19697--19705, 2023.

\bibitem[Barron et~al.(2024)Barron, Park, Mildenhall, Flynn, Verbin, Srinivasan, Hedman, Henzler, and Martin-Brualla]{campzipnerf2024}
Jonathan~T. Barron, Keunhong Park, Ben Mildenhall, John Flynn, Dor Verbin, Pratul Srinivasan, Peter Hedman, Philipp Henzler, and Ricardo Martin-Brualla.
\newblock {CamP Zip-NeRF}: {A} {Code} {Release} for {CamP} and {Zip-NeRF}, 2024.

\bibitem[Bemana et~al.(2022)Bemana, Myszkowski, Revall~Frisvad, Seidel, and Ritschel]{bemana2022eikonal}
Mojtaba Bemana, Karol Myszkowski, Jeppe Revall~Frisvad, Hans-Peter Seidel, and Tobias Ritschel.
\newblock {Eikonal Fields for Refractive Novel-view Synthesis}.
\newblock In \emph{ACM SIGGRAPH 2022 Conference Proceedings}, pages 1--9, 2022.

\bibitem[Bergen and Adelson(1991)]{bergen1991plenoptic}
James~R Bergen and Edward~H Adelson.
\newblock {The Plenoptic Function and The Elements of Early Vision}.
\newblock \emph{Computational Models of Visual Processing}, 1:\penalty0 8, 1991.

\bibitem[Boss et~al.(2021{\natexlab{a}})Boss, Braun, Jampani, Barron, Liu, and Lensch]{boss2021nerd}
Mark Boss, Raphael Braun, Varun Jampani, Jonathan~T Barron, Ce Liu, and Hendrik Lensch.
\newblock {{NeRD}: Neural Reflectance Decomposition from Image Collections}.
\newblock In \emph{Proceedings of the IEEE/CVF International Conference on Computer Vision}, pages 12684--12694, 2021{\natexlab{a}}.

\bibitem[Boss et~al.(2021{\natexlab{b}})Boss, Jampani, Braun, Liu, Barron, and Lensch]{boss2021neural}
Mark Boss, Varun Jampani, Raphael Braun, Ce Liu, Jonathan Barron, and Hendrik Lensch.
\newblock {{Neural-PIL}: Neural Pre-Integrated Lighting for Reflectance Decomposition}.
\newblock \emph{Advances in Neural Information Processing Systems}, 34:\penalty0 10691--10704, 2021{\natexlab{b}}.

\bibitem[Boss et~al.(2022)Boss, Engelhardt, Kar, Li, Sun, Barron, Lensch, and Jampani]{boss2022samurai}
Mark Boss, Andreas Engelhardt, Abhishek Kar, Yuanzhen Li, Deqing Sun, Jonathan~T. Barron, Hendrik Lensch, and Varun Jampani.
\newblock {{SAMURAI}: Shape And Material from Unconstrained Real-world Arbitrary Image collections}.
\newblock In \emph{Advances in Neural Information Processing Systems}, 2022.

\bibitem[Bradbury et~al.(2018)Bradbury, Frostig, Hawkins, Johnson, Leary, Maclaurin, Necula, Paszke, Vander{P}las, Wanderman-{M}ilne, and Zhang]{jax2018github}
James Bradbury, Roy Frostig, Peter Hawkins, Matthew~James Johnson, Chris Leary, Dougal Maclaurin, George Necula, Adam Paszke, Jake Vander{P}las, Skye Wanderman-{M}ilne, and Qiao Zhang.
\newblock {JAX}: composable transformations of {P}ython+{N}um{P}y programs, 2018.

\bibitem[Cheng et~al.(2022)Cheng, Li, Hartley, Zheng, and Sato]{cheng2022diffeomorphic}
Ziang Cheng, Hongdong Li, Richard Hartley, Yinqiang Zheng, and Imari Sato.
\newblock {Diffeomorphic Neural Surface Parameterization for 3D and Reflectance Acquisition}.
\newblock In \emph{ACM SIGGRAPH 2022 Conference Proceedings}, pages 1--10, 2022.

\bibitem[Conway and Sloane(2013)]{conway2013sphere}
John~Horton Conway and Neil James~Alexander Sloane.
\newblock \emph{Sphere packings, lattices and groups}.
\newblock Springer Science \& Business Media, 2013.

\bibitem[Dogaru et~al.(2023)Dogaru, Ardelean, Ignatyev, Zakharov, and Burnaev]{dogaru2022sphere}
Andreea Dogaru, Andrei-Timotei Ardelean, Savva Ignatyev, Egor Zakharov, and Evgeny Burnaev.
\newblock Sphere-guided training of neural implicit surfaces.
\newblock In \emph{Proceedings of the IEEE/CVF Conference on Computer Vision and Pattern Recognition (CVPR)}, pages 20844--20853, 2023.

\bibitem[Frisch and Hanebeck(2023)]{frisch2023vmf_sampling}
Daniel Frisch and Uwe~D. Hanebeck.
\newblock Deterministic {Von Mises–Fisher} sampling on the sphere using fibonacci lattices.
\newblock In \emph{2023 IEEE Symposium Sensor Data Fusion and International Conference on Multisensor Fusion and Integration (SDF-MFI)}, pages 1--8, 2023.

\bibitem[Fu et~al.(2022)Fu, Xu, Ong, and Tao]{fu2022geoneus}
Qiancheng Fu, Qingshan Xu, Yew-Soon Ong, and Wenbing Tao.
\newblock {{Geo-Neus}: Geometry-Consistent Neural Implicit Surfaces Learning for Multi-view Reconstruction}.
\newblock In \emph{Advances in Neural Information Processing Systems}, 2022.

\bibitem[Garces et~al.(2017)Garces, Echevarria, Zhang, Wu, Zhou, and Gutierrez]{garces2017intrinsic}
Elena Garces, Jose~I Echevarria, Wen Zhang, Hongzhi Wu, Kun Zhou, and Diego Gutierrez.
\newblock {Intrinsic Light Field Images}.
\newblock In \emph{Computer Graphics Forum}, pages 589--599. Wiley Online Library, 2017.

\bibitem[Garces et~al.(2022)Garces, Rodriguez-Pardo, Casas, and Lopez-Moreno]{garcesintrinsic2022}
Elena Garces, Carlos Rodriguez-Pardo, Dan Casas, and Jorge Lopez-Moreno.
\newblock {A Survey on Intrinsic Images: Delving Deep into {Lambert} and Beyond}.
\newblock \emph{Int. J. Comput. Vision}, 130\penalty0 (3):\penalty0 836–868, 2022.

\bibitem[Ge et~al.(2023)Ge, Hu, Zhao, Liu, and Chen]{ge2023ref}
Wenhang Ge, Tao Hu, Haoyu Zhao, Shu Liu, and Ying-Cong Chen.
\newblock {Ref-NeUS}: Ambiguity-reduced neural implicit surface learning for multi-view reconstruction with reflection.
\newblock In \emph{Proceedings of the IEEE/CVF International Conference on Computer Vision}, pages 4251--4260, 2023.

\bibitem[Govindarajan et~al.(2025)Govindarajan, Rebain, Yi, and Tagliasacchi]{govindarajan2025radiant}
Shrisudhan Govindarajan, Daniel Rebain, Kwang~Moo Yi, and Andrea Tagliasacchi.
\newblock Radiant foam: Real-time differentiable ray tracing.
\newblock \emph{arXiv preprint arXiv:2502.01157}, 2025.

\bibitem[Guo et~al.(2022)Guo, Kang, Bao, He, and Zhang]{guo2022nerfren}
Yuan-Chen Guo, Di Kang, Linchao Bao, Yu He, and Song-Hai Zhang.
\newblock {{NeRFReN}: Neural Radiance Fields with Reflections}.
\newblock In \emph{Proceedings of the IEEE/CVF Conference on Computer Vision and Pattern Recognition}, pages 18409--18418, 2022.

\bibitem[Huang et~al.(2024)Huang, Yu, Chen, Geiger, and Gao]{huang20242d}
Binbin Huang, Zehao Yu, Anpei Chen, Andreas Geiger, and Shenghua Gao.
\newblock 2d gaussian splatting for geometrically accurate radiance fields.
\newblock In \emph{ACM SIGGRAPH 2024 conference papers}, pages 1--11, 2024.

\bibitem[Irshad et~al.(2024)Irshad, Comi, Lin, Heppert, Valada, Ambrus, Kira, and Tremblay]{irshad2024neuralfieldsroboticssurvey}
Muhammad~Zubair Irshad, Mauro Comi, Yen-Chen Lin, Nick Heppert, Abhinav Valada, Rares Ambrus, Zsolt Kira, and Jonathan Tremblay.
\newblock Neural fields in robotics: A survey.
\newblock \emph{arXiv preprint arXiv:2410.20220}, 2024.

\bibitem[Jiang et~al.(2025)Jiang, Sivaram, Peng, and Ramamoorthi]{jiang2025geometry}
Kaiwen Jiang, Venkataram Sivaram, Cheng Peng, and Ravi Ramamoorthi.
\newblock Geometry field splatting with gaussian surfels.
\newblock In \emph{Proceedings of the Computer Vision and Pattern Recognition Conference}, pages 5752--5762, 2025.

\bibitem[Kerbl et~al.(2023)Kerbl, Kopanas, Leimk{\"u}hler, and Drettakis]{kerbl20233d}
Bernhard Kerbl, Georgios Kopanas, Thomas Leimk{\"u}hler, and George Drettakis.
\newblock 3d gaussian splatting for real-time radiance field rendering.
\newblock \emph{ACM Trans. Graph.}, 42\penalty0 (4):\penalty0 139--1, 2023.

\bibitem[Kim et~al.(2022)Kim, Seo, and Han]{kim2022infonerf}
Mijeong Kim, Seonguk Seo, and Bohyung Han.
\newblock {InfoNeRF}: Ray entropy minimization for few-shot neural volume rendering.
\newblock In \emph{Proceedings of the IEEE/CVF Conference on Computer Vision and Pattern Recognition}, pages 12912--12921, 2022.

\bibitem[Li et~al.(2024)Li, Lyu, Di, Zhai, Lee, and Tombari]{li2024geogaussian}
Yanyan Li, Chenyu Lyu, Yan Di, Guangyao Zhai, Gim~Hee Lee, and Federico Tombari.
\newblock Geogaussian: Geometry-aware gaussian splatting for scene rendering.
\newblock In \emph{European Conference on Computer Vision}, pages 441--457. Springer, 2024.

\bibitem[Li and Snavely(2018)]{Li_2018_ECCV}
Zhengqi Li and Noah Snavely.
\newblock {CGIntrinsics: Better Intrinsic Image Decomposition through Physically-Based Rendering}.
\newblock In \emph{Proceedings of the European Conference on Computer Vision (ECCV)}, 2018.

\bibitem[Liu et~al.(2025)Liu, Sun, Chen, Wang, and Feng]{liu2025deformable}
Rong Liu, Dylan Sun, Meida Chen, Yue Wang, and Andrew Feng.
\newblock Deformable beta splatting.
\newblock \emph{arXiv preprint arXiv:2501.18630}, 2025.

\bibitem[Ma et~al.(2023)Ma, Agrawal, Turki, Kim, Gao, Sander, Zollh{\"o}fer, and Richardt]{ma2023specnerf}
Li Ma, Vasu Agrawal, Haithem Turki, Changil Kim, Chen Gao, Pedro Sander, Michael Zollh{\"o}fer, and Christian Richardt.
\newblock {SpecNeRF}: Gaussian directional encoding for specular reflections.
\newblock \emph{arXiv preprint arXiv:2312.13102}, 2023.

\bibitem[Martin-Brualla et~al.(2021)Martin-Brualla, Radwan, Sajjadi, Barron, Dosovitskiy, and Duckworth]{martin2021nerf}
Ricardo Martin-Brualla, Noha Radwan, Mehdi~SM Sajjadi, Jonathan~T Barron, Alexey Dosovitskiy, and Daniel Duckworth.
\newblock {{NeRF} in the Wild: Neural Radiance Fields for Unconstrained Photo Collections}.
\newblock In \emph{Proceedings of the IEEE/CVF Conference on Computer Vision and Pattern Recognition}, pages 7210--7219, 2021.

\bibitem[Mildenhall et~al.(2020)Mildenhall, Srinivasan, Tancik, Barron, Ramamoorthi, and Ng]{mildenhall2020nerf}
Ben Mildenhall, Pratul~P. Srinivasan, Matthew Tancik, Jonathan~T. Barron, Ravi Ramamoorthi, and Ren Ng.
\newblock {{NeRF}: Representing Scenes as Neural Radiance Fields for View Synthesis}.
\newblock In \emph{ECCV}, 2020.

\bibitem[M\"uller et~al.(2022)M\"uller, Evans, Schied, and Keller]{mueller2022instant}
Thomas M\"uller, Alex Evans, Christoph Schied, and Alexander Keller.
\newblock Instant neural graphics primitives with a multiresolution hash encoding.
\newblock \emph{ACM Trans. Graph.}, 41\penalty0 (4):\penalty0 102:1--102:15, 2022.

\bibitem[Niemeyer et~al.(2022)Niemeyer, Barron, Mildenhall, Sajjadi, Geiger, and Radwan]{niemeyer2022regnerf}
Michael Niemeyer, Jonathan~T Barron, Ben Mildenhall, Mehdi~SM Sajjadi, Andreas Geiger, and Noha Radwan.
\newblock {{RegNeRF}: Regularizing Neural Radiance Fields for View Synthesis From Sparse Inputs}.
\newblock In \emph{Proceedings of the IEEE/CVF Conference on Computer Vision and Pattern Recognition}, pages 5480--5490, 2022.

\bibitem[Oechsle et~al.(2021)Oechsle, Peng, and Geiger]{oechsle2021unisurf}
Michael Oechsle, Songyou Peng, and Andreas Geiger.
\newblock {{UNISURF}: Unifying Neural Implicit Surfaces and Radiance Fields for Multi-View Reconstruction}.
\newblock In \emph{Proceedings of the IEEE/CVF International Conference on Computer Vision}, pages 5589--5599, 2021.

\bibitem[Rebain et~al.(2022)Rebain, Matthews, Yi, Lagun, and Tagliasacchi]{rebain2022lolnerf}
Daniel Rebain, Mark Matthews, Kwang~Moo Yi, Dmitry Lagun, and Andrea Tagliasacchi.
\newblock {{LolNeRF}: Learn From One Look}.
\newblock In \emph{Proceedings of the IEEE/CVF Conference on Computer Vision and Pattern Recognition}, pages 1558--1567, 2022.

\bibitem[Rosu and Behnke(2023)]{rosu2023permutosdf}
Radu~Alexandru Rosu and Sven Behnke.
\newblock Permutosdf: Fast multi-view reconstruction with implicit surfaces using permutohedral lattices.
\newblock In \emph{Proceedings of the IEEE/CVF Conference on Computer Vision and Pattern Recognition}, pages 8466--8475, 2023.

\bibitem[Sabour et~al.(2023)Sabour, Vora, Duckworth, Krasin, Fleet, and Tagliasacchi]{sabour2023robustnerf}
Sara Sabour, Suhani Vora, Daniel Duckworth, Ivan Krasin, David~J. Fleet, and Andrea Tagliasacchi.
\newblock Robustnerf: Ignoring distractors with robust losses.
\newblock In \emph{2023 IEEE/CVF Conference on Computer Vision and Pattern Recognition (CVPR)}, pages 20626--20636, 2023.

\bibitem[Somraj et~al.(2023)Somraj, Karanayil, and Soundararajan]{nagabushan2023siggraphasia}
Nagabhushan Somraj, Adithyan Karanayil, and Rajiv Soundararajan.
\newblock {SimpleNeRF}: Regularizing sparse input neural radiance fields with simpler solutions.
\newblock In \emph{SIGGRAPH Asia 2023 Conference Papers}, New York, NY, USA, 2023. Association for Computing Machinery.

\bibitem[Tancik et~al.(2022)Tancik, Casser, Yan, Pradhan, Mildenhall, Srinivasan, Barron, and Kretzschmar]{tancik2022block}
Matthew Tancik, Vincent Casser, Xinchen Yan, Sabeek Pradhan, Ben Mildenhall, Pratul~P Srinivasan, Jonathan~T Barron, and Henrik Kretzschmar.
\newblock {Block-{NeRF}: Scalable Large Scene Neural View Synthesis}.
\newblock In \emph{Proceedings of the IEEE/CVF Conference on Computer Vision and Pattern Recognition}, pages 8248--8258, 2022.

\bibitem[Verbin et~al.(2022)Verbin, Hedman, Mildenhall, Zickler, Barron, and Srinivasan]{verbin2022ref}
Dor Verbin, Peter Hedman, Ben Mildenhall, Todd Zickler, Jonathan~T Barron, and Pratul~P Srinivasan.
\newblock {Ref-{NeRF}: Structured View-Dependent Appearance For Neural Radiance Fields}.
\newblock In \emph{2022 IEEE/CVF Conference on Computer Vision and Pattern Recognition (CVPR)}, pages 5481--5490. IEEE, 2022.

\bibitem[Verbin et~al.(2024)Verbin, Srinivasan, Hedman, Mildenhall, Attal, Szeliski, and Barron]{verbin2024nerf}
Dor Verbin, Pratul~P Srinivasan, Peter Hedman, Ben Mildenhall, Benjamin Attal, Richard Szeliski, and Jonathan~T Barron.
\newblock {NeRF-Casting}: Improved view-dependent appearance with consistent reflections.
\newblock \emph{arXiv preprint arXiv:2405.14871}, 2024.

\bibitem[Wang et~al.(2022)Wang, Han, Habermann, Daniilidis, Theobalt, and Liu]{wang2022neus2}
Yiming Wang, Qin Han, Marc Habermann, Kostas Daniilidis, Christian Theobalt, and Lingjie Liu.
\newblock {{NeuS2}: Fast Learning of Neural Implicit Surfaces for Multi-view Reconstruction}.
\newblock \emph{arXiv preprint arXiv:2212.05231}, 2022.

\bibitem[Wang et~al.(2024)Wang, Wu, Zhang, and Xu]{wang2024learning}
Yuxin Wang, Qianyi Wu, Guofeng Zhang, and Dan Xu.
\newblock Learning 3d geometry and feature consistent gaussian splatting for object removal.
\newblock In \emph{European Conference on Computer Vision}, pages 1--17. Springer, 2024.

\bibitem[Wei et~al.(2024)Wei, Wu, Zheng, Rezatofighi, and Cai]{wei2024normal}
Meng Wei, Qianyi Wu, Jianmin Zheng, Hamid Rezatofighi, and Jianfei Cai.
\newblock Normal-gs: 3d gaussian splatting with normal-involved rendering.
\newblock In \emph{The Thirty-eighth Annual Conference on Neural Information Processing Systems}, 2024.

\bibitem[Wood et~al.(2000)Wood, Azuma, Aldinger, Curless, Duchamp, Salesin, and Stuetzle]{woodsurface2000}
Daniel~N. Wood, Daniel~I. Azuma, Ken Aldinger, Brian Curless, Tom Duchamp, David~H. Salesin, and Werner Stuetzle.
\newblock {Surface Light Fields for {3D} Photography}.
\newblock In \emph{Proceedings of the 27th Annual Conference on Computer Graphics and Interactive Techniques}, page 287–296, USA, 2000. ACM Press/Addison-Wesley Publishing Co.

\bibitem[Yang et~al.(2022)Yang, Bao, Zeng, Bao, Zhang, Cui, and Zhang]{yang2022neumesh}
Bangbang Yang, Chong Bao, Junyi Zeng, Hujun Bao, Yinda Zhang, Zhaopeng Cui, and Guofeng Zhang.
\newblock {Neumesh: Learning Disentangled Neural Mesh-Based Implicit Field for Geometry and Texture Editing}.
\newblock In \emph{Computer Vision--ECCV 2022: 17th European Conference, Tel Aviv, Israel, October 23--27, 2022, Proceedings, Part XVI}, pages 597--614. Springer, 2022.

\bibitem[Yao et~al.(2022)Yao, Zhang, Liu, Qu, Fang, McKinnon, Tsin, and Quan]{yao2022neilf}
Yao Yao, Jingyang Zhang, Jingbo Liu, Yihang Qu, Tian Fang, David McKinnon, Yanghai Tsin, and Long Quan.
\newblock {{NeILF}: Neural Incident Light Field for Physically-Based Material Estimation}.
\newblock In \emph{Computer Vision--ECCV 2022: 17th European Conference, Tel Aviv, Israel, October 23--27, 2022, Proceedings, Part XXXI}, pages 700--716. Springer, 2022.

\bibitem[Yariv et~al.(2021)Yariv, Gu, Kasten, and Lipman]{yariv2021volume}
Lior Yariv, Jiatao Gu, Yoni Kasten, and Yaron Lipman.
\newblock {Volume Rendering of Neural Implicit Surfaces}.
\newblock \emph{Advances in Neural Information Processing Systems}, 34:\penalty0 4805--4815, 2021.

\bibitem[Yu et~al.(2022)Yu, Peng, Niemeyer, Sattler, and Geiger]{yu2022monosdf}
Zehao Yu, Songyou Peng, Michael Niemeyer, Torsten Sattler, and Andreas Geiger.
\newblock {Mono{SDF}: Exploring Monocular Geometric Cues for Neural Implicit Surface Reconstruction}.
\newblock In \emph{Advances in Neural Information Processing Systems}, 2022.

\bibitem[Zhang et~al.(2021{\natexlab{a}})Zhang, Yang, Tulsiani, and Ramanan]{zhang2021ners}
Jason Zhang, Gengshan Yang, Shubham Tulsiani, and Deva Ramanan.
\newblock {{NeRS}: Neural Reflectance Surfaces for Sparse-view 3D Reconstruction in the Wild}.
\newblock \emph{Advances in Neural Information Processing Systems}, 34:\penalty0 29835--29847, 2021{\natexlab{a}}.

\bibitem[Zhang et~al.(2022)Zhang, Luan, Li, and Snavely]{zhang2022iron}
Kai Zhang, Fujun Luan, Zhengqi Li, and Noah Snavely.
\newblock {{IRON}: Inverse Rendering by Optimizing Neural {SDFs} and Materials from Photometric Images}.
\newblock In \emph{Proceedings of the IEEE/CVF Conference on Computer Vision and Pattern Recognition}, pages 5565--5574, 2022.

\bibitem[Zhang et~al.(2021{\natexlab{b}})Zhang, Srinivasan, Deng, Debevec, Freeman, and Barron]{zhang2021nerfactor}
Xiuming Zhang, Pratul~P Srinivasan, Boyang Deng, Paul Debevec, William~T Freeman, and Jonathan~T Barron.
\newblock {Nerfactor: Neural Factorization of Shape and Reflectance Under an Unknown Illumination}.
\newblock \emph{ACM Transactions on Graphics (TOG)}, 40\penalty0 (6):\penalty0 1--18, 2021{\natexlab{b}}.

\bibitem[Zhu et~al.(2023)Zhu, Yang, Wang, Zheng, and Guibas]{Zhu_2023_CVPR}
Bingfan Zhu, Yanchao Yang, Xulong Wang, Youyi Zheng, and Leonidas Guibas.
\newblock {VDN-NeRF}: Resolving shape-radiance ambiguity via view-dependence normalization.
\newblock In \emph{Proceedings of the IEEE/CVF Conference on Computer Vision and Pattern Recognition (CVPR)}, pages 35--45, 2023.

\end{thebibliography}
}

\end{document}

\maketitle
\section{Deterministic Uniform Sphere Sampling}
Sampling over spherical domains requires careful construction to avoid over-sampling or under-sampling, particularly as the azimuth coordinate $\theta \rightarrow \pm \frac{\pi}{2}$. We experimented extensively, and found that traditional uniform sampling leads to unintended clusters of sampling directions which led to artefacts in the specular term. This can be seen in the random sampling row of Table~\ref{tab:ablation} in the main paper. For this reason, we adapt the approach proposed by Frisch et al.~\cite{frisch2023vmf_sampling} for a von Mises-Fisher (vMF) distribution in $\mathbb{S}^2$ to provide a deterministic uniform sample of a sphere.

The vMF distribution,
\begin{equation}
    f(\vb{x})=\frac{\kappa}{2\pi(e^\kappa-e^{-\kappa})}e^{\kappa\pmb{\mu}^\top\vb{x}},
\end{equation}
is a probability distribution with mean $\pmb{\mu}\in\mathbb{S}^2$ and concentration parameter $\kappa\geq0$ on the 2-sphere. In the limit as $\kappa\to0$, $f(\vb{x})=(4\pi)^{-1}$, resulting in a uniform distribution.

Frisch et al.~\cite{frisch2023vmf_sampling} sample this distribution using a Fibonacci-Kronecker lattice, with coordinates $\vb{x}_{FK}\in\mathbb{S}^2$,
\begin{equation}
    \vb{x}_{F,i} = \mqty[w \\
                        \sqrt{1-w^2}\cos{\frac{2\pi i}{\Phi}}\\
                        \sqrt{1-w^2}\sin{\frac{2\pi i}{\Phi}}],\; i\in\{0,\ldots,N-1\}
\end{equation}
where $\frac{1}{\Phi}=\frac{\sqrt{5}-1}{2}$ is the Golden Ratio, and
\begin{equation}
    w = 1 + \frac{1}{\kappa}\mathrm{log1p}\qty(\frac{2i-1}{2N}\mathrm{expm1}(-2\kappa)),
\end{equation}
for $\mathrm{log1p}(x)=\log(1+x)$ and $\mathrm{expm1}(x)=\exp(x)-1$. Since we want a uniform distribution, we take the limit $\kappa \rightarrow 0$, resulting in,
\begin{equation}
    \lim_{\kappa \rightarrow 0} w = w_{\kappa=0} = \frac{1-2i+N}{N}.
\end{equation}
We wish to maintain the distribution of directions of the sphere, but partition the samples through the volume of the unit ball. We introduce a discrete radius for each sample,
\begin{equation}
    \vb{x}_i = \frac{1}{\log_2{N}}(1 + i\bmod{\log_2{N}})\vb{x}_{F,i},
\end{equation}
to partition the samples into $\log_2 N$ shells in the unit ball.

\begin{figure*}
    \centering

        \begin{subfigure}[t]{0.48\textwidth}
        \centering
        \includegraphics[width=\textwidth]{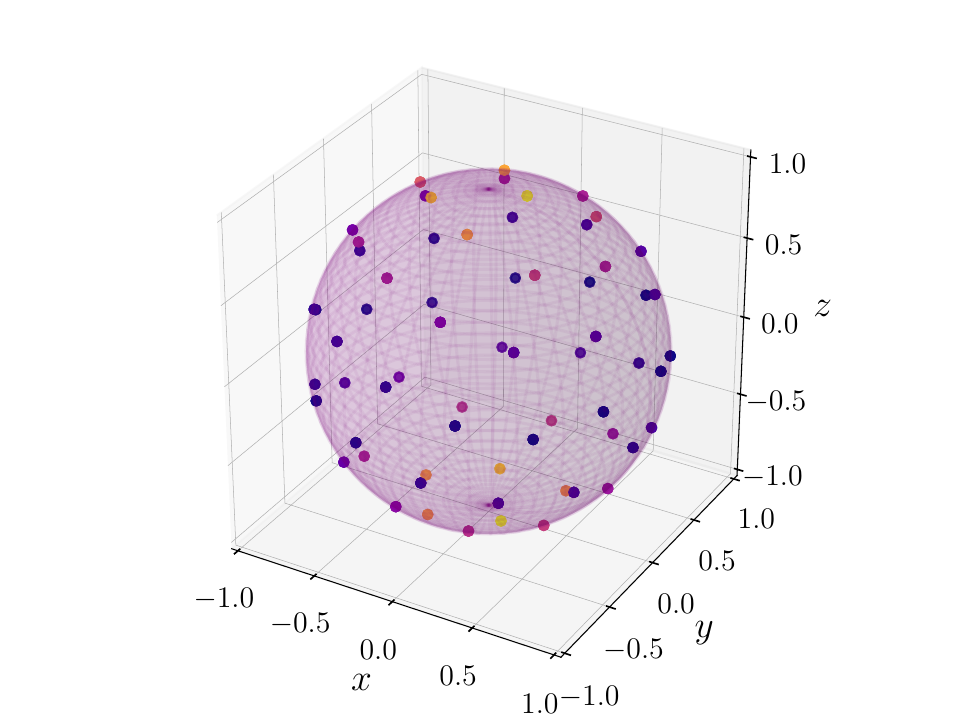}
        \caption{3D plot of deterministic samples on the sphere before applying the $\log_2 N$ radii partitions to sample the unit ball.}
    \end{subfigure}
    \quad
    \begin{subfigure}[t]{0.48\textwidth}
        \centering
        \includegraphics[width=\textwidth]{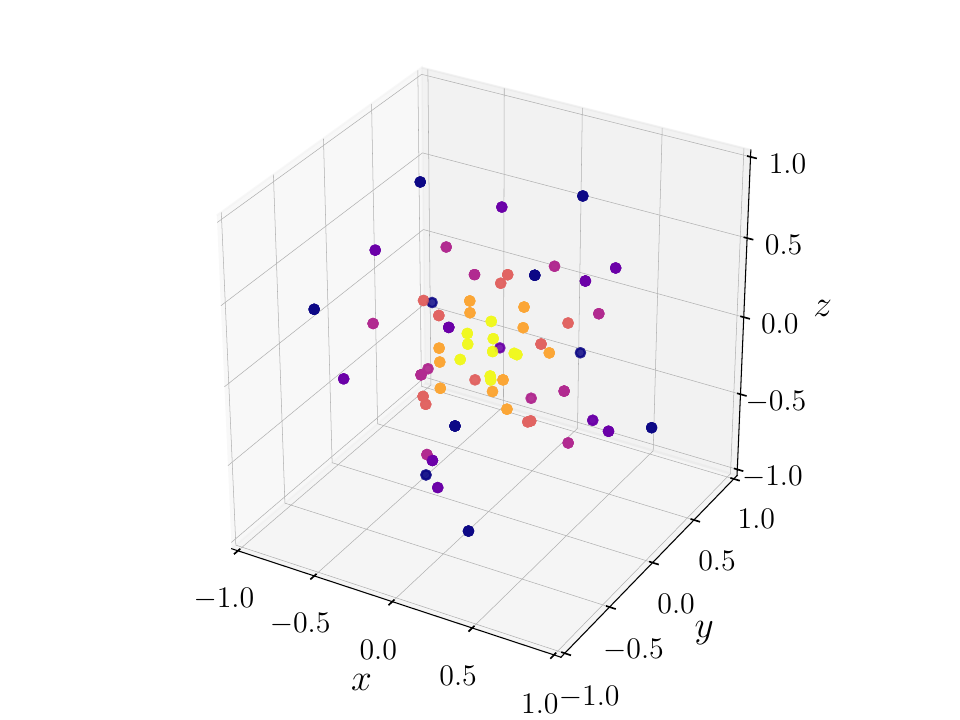}
        \caption{3D plot of deterministic sampling. These point locations are used in the proposed spatial sampling batch.}
    \end{subfigure}
\quad
    \begin{subfigure}[t]{0.48\textwidth}
        \centering
        \includegraphics[width=\textwidth]{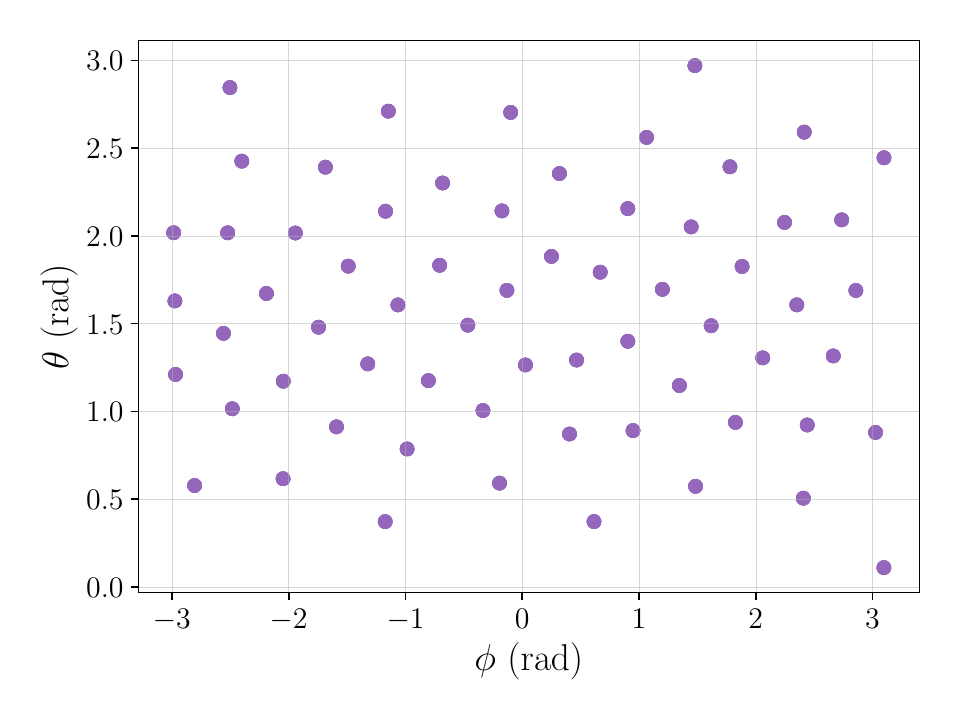}
        \caption{Directions through which we sample. We see that the lattice uniformly distributes sample directions, taking into account the warping close to the poles.}
    \end{subfigure}
    \quad
            \begin{subfigure}[t]{0.48\textwidth}
        \centering
        \includegraphics[width=\textwidth]{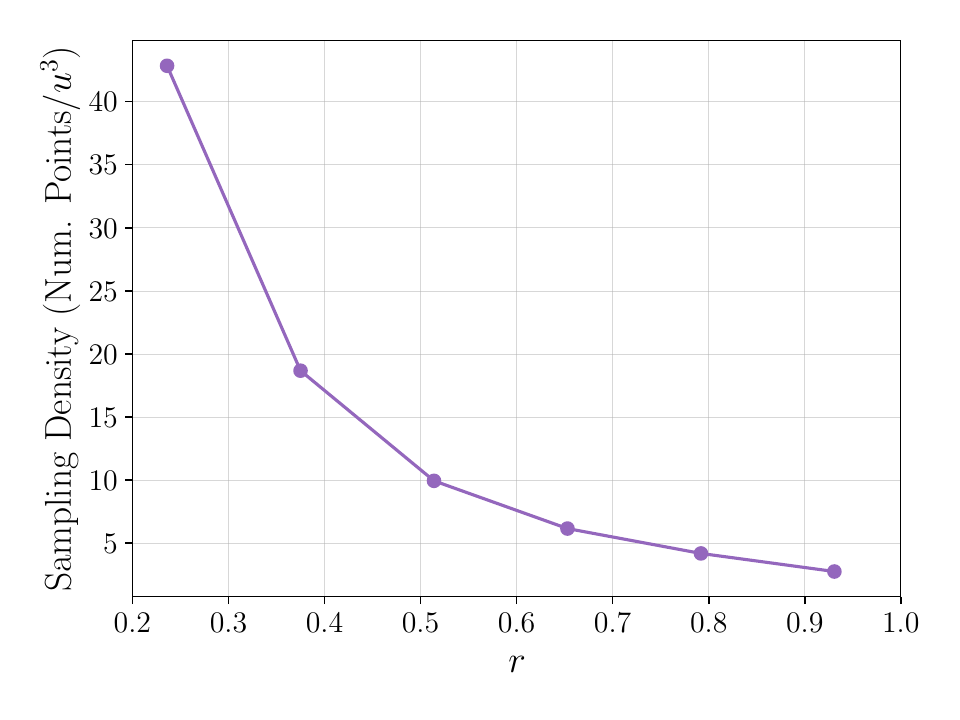}
        \caption{The density (number of points per unit volume) of each shell in the proposed spatial sampling batch. Samples closest to the surface have a higher density and therefore greater effect during regularisation.}
    \end{subfigure}
    \caption{Example directional batch and its polar coordinate map, indicating samples used for regularisation and volumetric density of sampling.}
    \label{fig:sample_structure_spatial}
\end{figure*}

We recognise that these shells, whilst uniform in direction have decreasing density away from the centre of the sphere. This helps in our use case for the sampling, with a higher density of samples closer to the centre of the ball (i.e. the surface). To avoid any bias which may arise in the volumetric samples, we generate a rotation matrix which is uniformly distributed in $\mathrm{SO}(3)$, using~\cite{avro1992fast}. Figure~\ref{fig:sample_structure_spatial} provides a visualisation of the direction and spatial coordinates for each sample using this approach.

% Some approaches utilise rejection sampling to ensure uniformity, however this is intractable for large numbers of samples. We utilise a uniform sampling which preserves uniformity around the poles of the sphere.
% Given random variables $u, v\sim \mathcal{U}\{0, 1\}$ drawn from a uniform distribution we derive angles,
% \begin{subequations}
%     \begin{equation}
%         \theta = \cos^{-1} (1-2u),
%     \end{equation}
%     \begin{equation}
%         \phi = 2\pi v,
%     \end{equation}
% \end{subequations}
% in the polar domain $\mathcal{P}$. In sampling the azimuth $\theta$ in this way, the spatial distribution of points as $\theta \rightarrow \pm \frac{\pi}{2}$ is uniform after mapping to the surface of a sphere: $\mathcal{P}\mapsto\mathbf{S}^2\in \mathbb{R}^3$.

% The points $\vb{p}\in \mathbf{S}^2$ are calculated in the usual way for a spherical to cartesian coordinate change,
% \begin{subequations}
%     \begin{equation}
%         x = \cos{\phi}\sin{\theta},
%     \end{equation}
%     \begin{equation}
%         y = \sin{\phi}\sin{\theta},
%     \end{equation}
%     \begin{equation}
%         z = \cos{\theta}.
%     \end{equation}
% \end{subequations}
We initialise a sphere of $N$ (in our case, $N=32$) samples prior to the training loop, and at each iteration generate a unique rotation matrix for each ray which is applied to this sphere to provide the unique sampling. Rays which view the same part of the scene are therefore unlikely to sample to same positions or directions during training.

% In Table~\ref{tab:determ} we present a short study comparing random uniform sampling of points on the sphere to our proposed deterministic sampling. Our deterministic sampling scheme achieves approximately a 13.2\% improvement to normals compared to a 5\% improvement attained by uniform samples on the sphere. The degradation in PSNR and disparity can be attributed to clustering of sample points leading to large regions which are sparsely regularised during training, and regions which are more heavily regularised.

% \begin{table}[htbp]
% \centering
% \caption{Deterministic sampling scheme study on the car scene from \textit{Shiny Objects}~\cite{verbin2022ref} dataset. Deterministic samples improve in all aspects by preventing clustering with our comparatively small number of additional samples in a regularisation batch.}
% \scriptsize
% \begin{tabular}{lcccc}
% \hline
% Method                                   & PSNR & SSIM & MAE & Disp. RMSE \\\hline
% Zip+Ref-NeRF                 & 30.30 & 0.957 & 15.167 & 0.209 \\
% \hline
% Ours, Uniform Random & 29.67 & 0.953 & 14.668 & 0.216 \\
% Ours, Deterministic         & \textbf{29.94} & \textbf{0.954} & \textbf{13.165} & \textbf{0.198} \\\hline
% \end{tabular}\label{tab:determ}
% \end{table}
\section{Surface Sampling Behaviour as Samples Converge}
During training of a NeRF with integrated positional encoding~\cite{barron2021mipnerf,barron2022mip} (IPE), frequencies of features increase as ray samples converge to a final value and the modelled Gaussian regions of space used for rendering reach minimum volumes. The proposed Surf-NeRF sampling scheme uses parameters from the integrated positional encoding to adapt the local region used for regularisation depending on the distribution of density along a ray. Localised distributions of density are regularised at finer scales, more strongly producing our intended effects later during training than earlier when the NeRF is still converging.

\begin{figure}
    \centering
    \includegraphics[width=\linewidth]{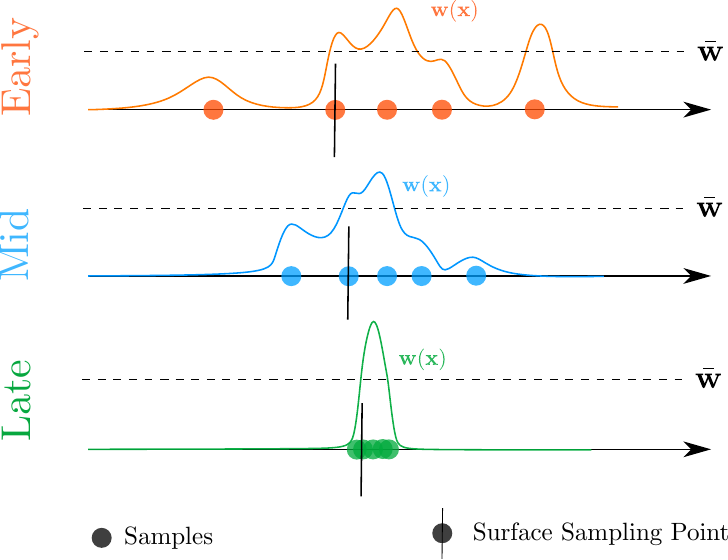}
    \caption{The evolution of weights as the scene representation converges during training. Surf-NeRF uses a first surface assumption to determine the location of sampling, which is robust to outliers along the ray.}
    \label{fig:sample_evol}
\end{figure}

Where the proposed sampling for the fine NeRF model in MipNeRF360~\cite{barron2022mip} produces highly clustered values, the spacing between subsequent samples becomes small; that is $\Delta~t\to~0$. The ideal surface representation within a NeRF is in fact a delta function along the ray, where density values are very large and highly clustered. Earlier during training, this distribution is more sparse. The evolution of the sample spacing is provided in Figure~\ref{fig:sample_evol}.

In this limit of convergence to a surface, the modelled Gaussians from IPE along a ray reach a minimum. Specifically, the sampling width $t_\delta$ between subsequent samples on the ray $t_i$ and $t_{i+1}$ approaches 0, as the mean sample approaches $t_i$,
\begin{subequations}
\begin{equation}
    t_\mu = \lim_{t_{i+1}\to t_i} \frac{t_{i+1}+t_i}{2} = t_i,
\end{equation}
\begin{equation}
    t_\delta = \lim_{t_{i+1}\to t_i} \frac{t_{i+1}-t_i}{2} = 0.
\end{equation}
\end{subequations}
Evaluating the equations for mean position $\mu_t$, variance $\sigma_t$ along the ray, and radial variance $\sigma_r$ in MipNeRF~\cite{barron2021mipnerf} we obtain,
\begin{subequations}
    \begin{equation}
        \lim_{t_{i+1}\to t_i} \mu_t = t_\mu,
    \end{equation}
        \begin{equation}
        \lim_{t_{i+1}\to t_i} \sigma^2_t = 0,
    \end{equation}
    \begin{equation}
        \lim_{t_{i+1}\to t_i} \sigma^2_r = \frac{\dot{r}^2t_\mu^2}{4},
    \end{equation}
\end{subequations}
where $\dot{r}$ is the radius of the modelled view cone of a ray at the image plane. The samples turn from 3D anisotropic Gaussians, to planar isotropic Gaussians in the scene orientated along the ray, thereby sampling in the radial direction only.

In Surf-NeRF, we use the radial standard deviation $\sigma_r$ to scale our sampling sphere. As a result, our additional sampling scales to the certainty of a surface being present. When the samples are placed far apart as the NeRF is still converging to localised density along a ray, $t_\delta$ is nonzero and $\sigma_r$ is greater than $\dot{r}t/2$. Regularisation therefore occurs over a larger area, encouraging geometric smoothness and normal consistency to be based on samples further away, adding density, but ablating finer details.
As training progresses, $\sigma_r$ approachs a minimum, ensuring sampling occurs within the bounds of a single pixel thereby retaining fine details and finetuning the representation.

\section{Lambertian Bias}
\begin{figure}
    \centering
    \includegraphics[width=\linewidth]{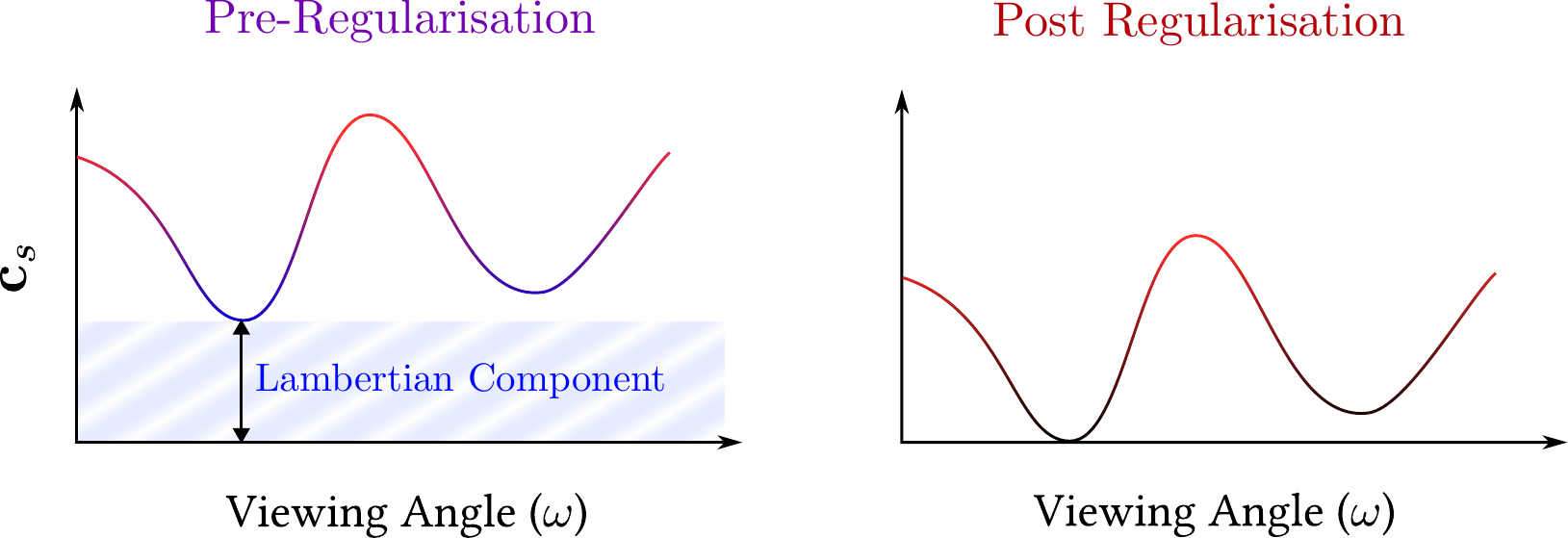}
    \caption{Our motivation in regularising the view-dependent appearance for geometry within a NeRF. The specular term absorbs the Lambertian colour, increasing reliance on floaters and false geometry. Removing this Lambertian bias forces the model to place density in a more physically consistent manner, reducing reliance on floaters.}
    \label{fig:specular_bias}
\end{figure}

Positional encoding helps MLPs to learn higher frequency information~\cite{mildenhall2020nerf} and reflection encodings~\cite{ma2023specnerf,verbin2022ref,verbin2024nerf} encourage a separation of low-frequency colour information with view-dependent components. However, we find in practice that the view-dependent term maintains a view-independent component across viewing angles. This is visualised in Figure~\ref{fig:specular_bias}, and in Figure~\ref{fig:appearance_improvements} in the main paper.

Early in training this phenomenon can be useful, allowing for density to accumulate independent of diffuse colour, and for surfaces with high roughness to develop low frequency specular distributions. In many cases, these low frequency components remain in the view-dependent appearance of the scene given minimal photometric error and no loss regularising this component. This has the drawback however that the representation does not need to assign a constant view-independent colour to points in the scene allowing for ambiguous placement of scene geometry. For most use cases involving novel view synthesis, this is not a critical consideration. 

However, by encouraging a unique Lambertian colour for each point in the scene where possible, we are able to better represent geometry by encouraging a view-independent appearance that must be consistent with all other observations of the scene. In essence, our regularisation minimises the magnitude of the specular term with the assumption that the view-dependent appearance should be sparse through viewing angles.
\section{Total Variation of Unordered Samples on a Sphere}
\begin{figure}[t!]
        \centering
        \includegraphics[width=0.5\textwidth]{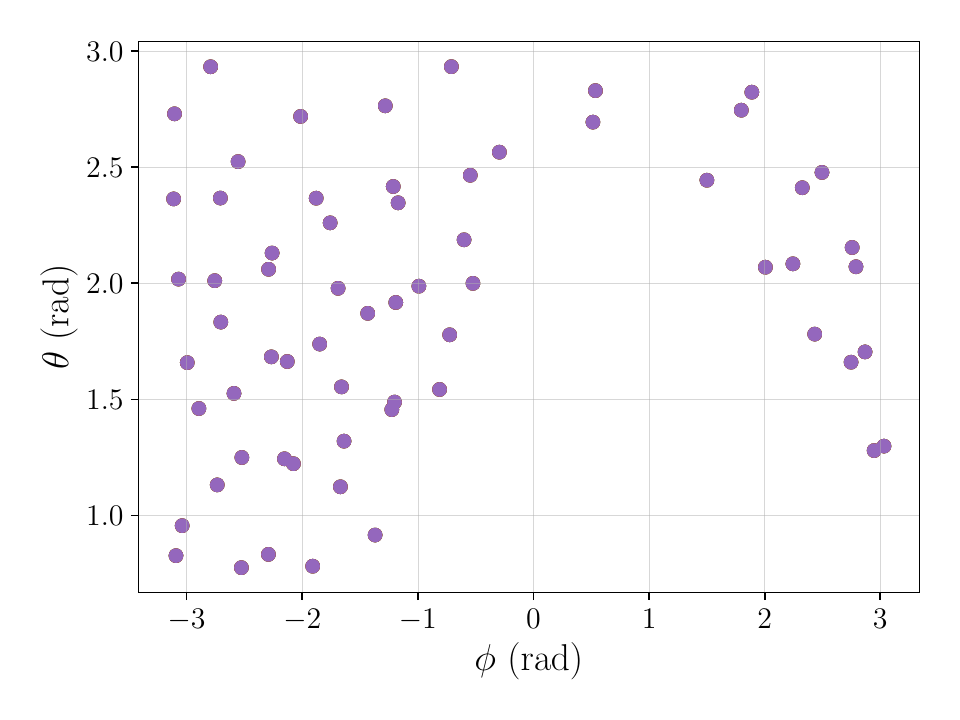}

    \caption{Example directional polar coordinate map. The spliced forward and backward facing directions creates two interleaved  sampling patterns, which is no longer regular. To address this, we use a graph total variation adapted for locations on the sphere accounting for irregularities.}
    \label{fig:total_var}
\end{figure}

We apply total variation regularisation $\mathcal{L}_s$ on the specular colour $\vb{c}_s$ to produce more smoothly varying view-dependent appearances. Using the normal of the surface, we produce samples over directions which are in the hemisphere visible to cameras in the scene and formulate total variation over this visible set of point. An example of this hemisphere is shown in Figure~\ref{fig:total_var}. 

To minimise total variation in specular colour, we select the $k$-nearest neighbours of directions on the hemisphere by angular distance. Selecting nearest neighbours produces the unique property that the regularisation is able to wrap around angles on the hemisphere. In essence, this is graph total variation over a subset of a hemispherical domain. Our implementation in JAX~\cite{jax2018github} allows gradients to propagate back to each sample, but not its neighbours thereby producing unique regularisation effects at each point.

We experimented with different numbers of $k$ when formulating this regularisation term. We found minimal improvement in geometric quality and marginally reduced visual fidelity going to higher numbers of $k$ above 3 most likely due to the small number of regularisation samples used and their comparative sparsity on the hemispherical domain.

\section{Model Details}
\begin{figure*}[htbp]
    \centering
    \includegraphics[width=0.8\textwidth]{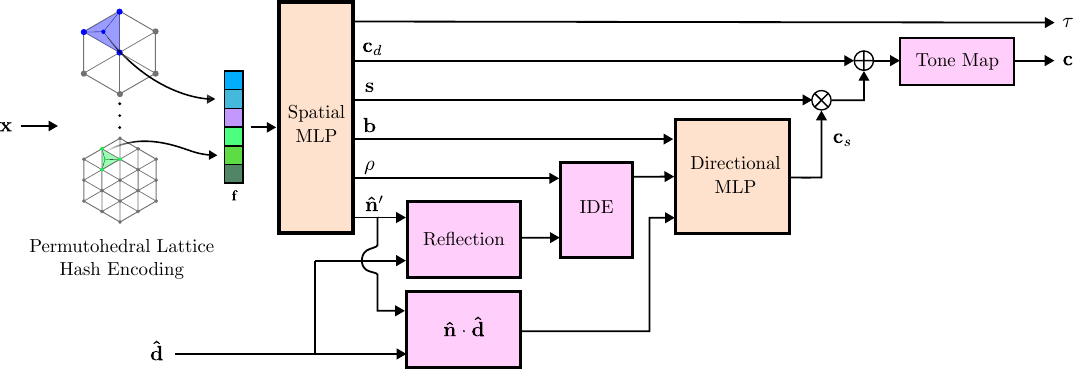}
    \caption{Visual depiction of the adapted Surf-NeRF model structure. We replace the cubic multi-resolution hash encoding in ZipNeRF~\cite{Barron_2023_ICCV} with a multi-resolution permutohedral lattice structure, similar to PermutoSDF~\cite{rosu2023permutosdf}. This is fused with the reflection encoding of Ref-NeRF~\cite{verbin2022ref}. We use the intermediate results for diffuse $\vb{c}_d$ and specular $\vb{c}_s$ colour terms to formulate the Surf-NeRF regularisation terms in terms of a Lambertian and specular scene appearance. Figure elements adapted from~\cite{rosu2023permutosdf, verbin2022ref}.}
    \label{fig:ziprefnerf}
\end{figure*}

In Figure~\ref{fig:ziprefnerf}, we provide a visual depiction of our model structure incorporating the multi-resolution hash encoding from PermutoSDF~\cite{rosu2023permutosdf} over the cubic grid in ZipNeRF~\cite{Barron_2023_ICCV} and the physics-inspired structure of Ref-NeRF~\cite{verbin2022ref}. For our positionally encoded implementation, we use the same model parameters as those in the original Ref-NeRF paper.

From the base ZipNeRF implementation, we alter the hash encoding to use a permutohedral lattice hash grid. Our model maintains the two proposal and final NeRF networks at the same resolution as ZipNeRF. We retain the 512, 2048 and finally 8192 grid sizes for our lattice in that order. We also maintain the proposed sampling strategy of the NeRF from ZipNeRF (64 samples for each proposal network, and 32 for the final NeRF). Our experimentation did not indicate substantial changes to the final quality of the NeRF with the same 128 final samples used in Ref-NeRF. In particular, in our first surface assumption for Surf-NeRF we note that the majority of the radiance along a ray should be the result of only a handful of samples clustered at the surface. Using a small number of final samples is in keeping with this assumption.

For the spatial MLP we use two layers with 128 neurons in each layer, a deeper network compared to base ZipNeRF helping the NeRF to learn the additional channels under the Ref-NeRF parameterisation. We use a 3-layer, 256 neuron wide MLP to learn the specular colour similar to ZipNeRF. The respective components from Ref-NeRF, namely the integrated directional encoding (IDE), tone mapping reflection calculation and dot-product remain unchanged in their construction and addition to the network.  

All weightings for losses in ZipNeRF~\cite{Barron_2023_ICCV} and Ref-NeRF~\cite{verbin2022ref} remain unchanged from previously published values. The Surf-NeRF regularisation terms have the following $\lambda$ weightings: our density smoothness $\mathcal{L}_d$ and normal consistency $\mathcal{L}_n$ terms have weightings $10^{-1}$, our specular bias term $\mathcal{L}_b$ has weighting $3\cdot10^{-2}$ and our specular total variation term $\mathcal{L}_s$ has weighting $10^{-3}$.

Our hash based implementations were trained for 25,000 iterations using a batch size of $2^{16}$, adjusted for GPU memory using the linear scaling rule. Our positionally encoded implementations were trained for 250,000 iterations using a batch size of $2^{14}$ rays. Optimisation followed the same learning schedule as ZipNeRF~\cite{Barron_2023_ICCV} and MipNeRF360~\cite{barron2022mip} for the hash-based and positionally encoded based implementations respectively.
\section{Per-Scene Results}
In this section we present the per-scene results of our baseline and proposed methods across the \textit{Shiny Objects}, \textit{Shiny Real}, and \textit{Koala} datasets. For the rendered \textit{Shiny Objects} dataset we provide bold values demonstrating the improved geometric performance attained from Surf-NeRF in comparison to other baseline methods across both the traditional positional encoded variants and the hash based methods. For the real datasets we demonstrate comparable visual performance to the existing methods with minimal degradation compared to the Ref-NeRF baselines we benchmark against.

\begin{figure*}
    \centering
    \includegraphics[width=\textwidth]{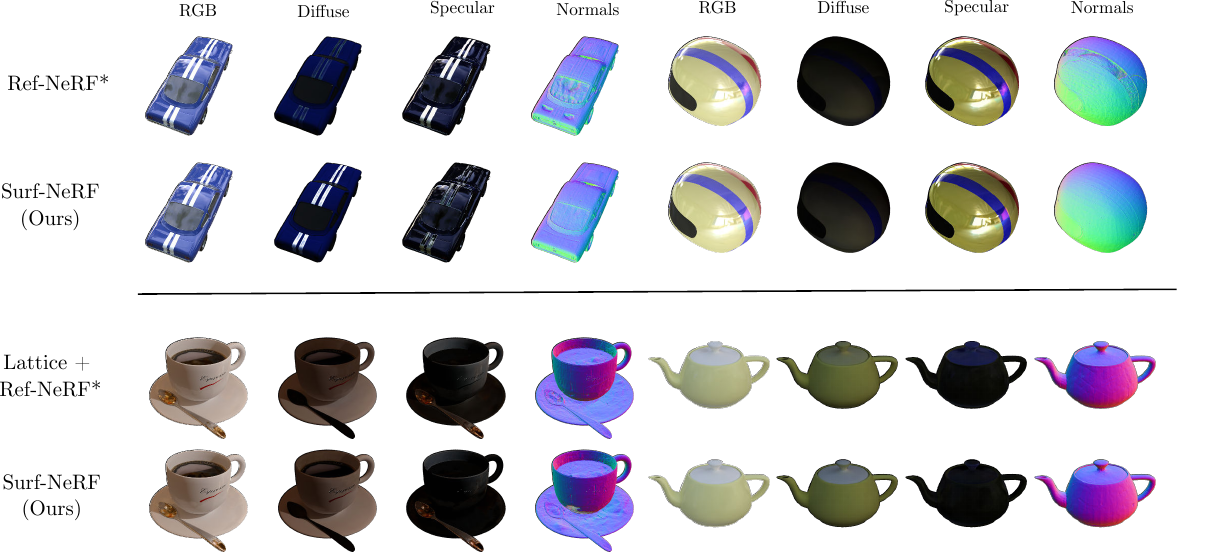}
    \caption{Examples from the \textit{Shiny Objects} dataset~\cite{verbin2022ref} for the hash based variants. Diffuse components like the white teacup surface and teapot lid are correctly removed from the specular term of the rendering, and normals show more complete surface structure. Ref-NeRF* indicates the Zip+Ref-NeRF baseline.}
    \label{fig:shiny_objects_supp}
\end{figure*}

\begin{figure}
    \centering
    \includegraphics[width=\linewidth]{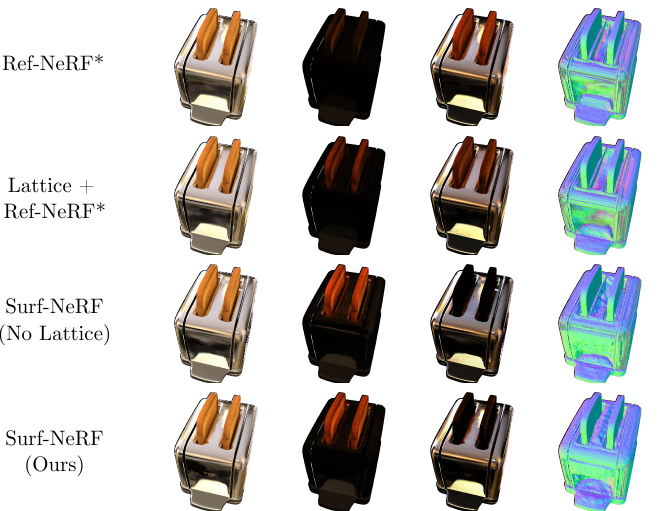}
    \caption{Comparison across all reflection based baseline models and the proposed method on the Toaster scene. Applying the proposed regularisation dramatically improves the separation of diffuse and specular scene content, and improved surface normals over a baseline reflection parameterisation. Some elements with little view dependence like interreflection of close scene content remains a challenge to shape radiance ambiguity.}
    \label{fig:toaster_scene}
\end{figure}

Table~\ref{tab:perscene_objects} outlines results for the \textit{Shiny Objects}~\cite{verbin2022ref} dataset. We show improved normals using our method across the positional encoded and grid-based approaches, and improved disparity RMSE scores. Figure~\ref{fig:shiny_objects_supp} provides additional qualitative results on this dataset showing separation of the Lambertian scene content and qualitatively improved geometry across all datasets. Figure~\ref{fig:toaster_scene} compares to all grid based baselines, showing improved geometry but remaining challenges with interreflection.

\begin{figure}[htbp]
    \centering
    \includegraphics[width=\linewidth]{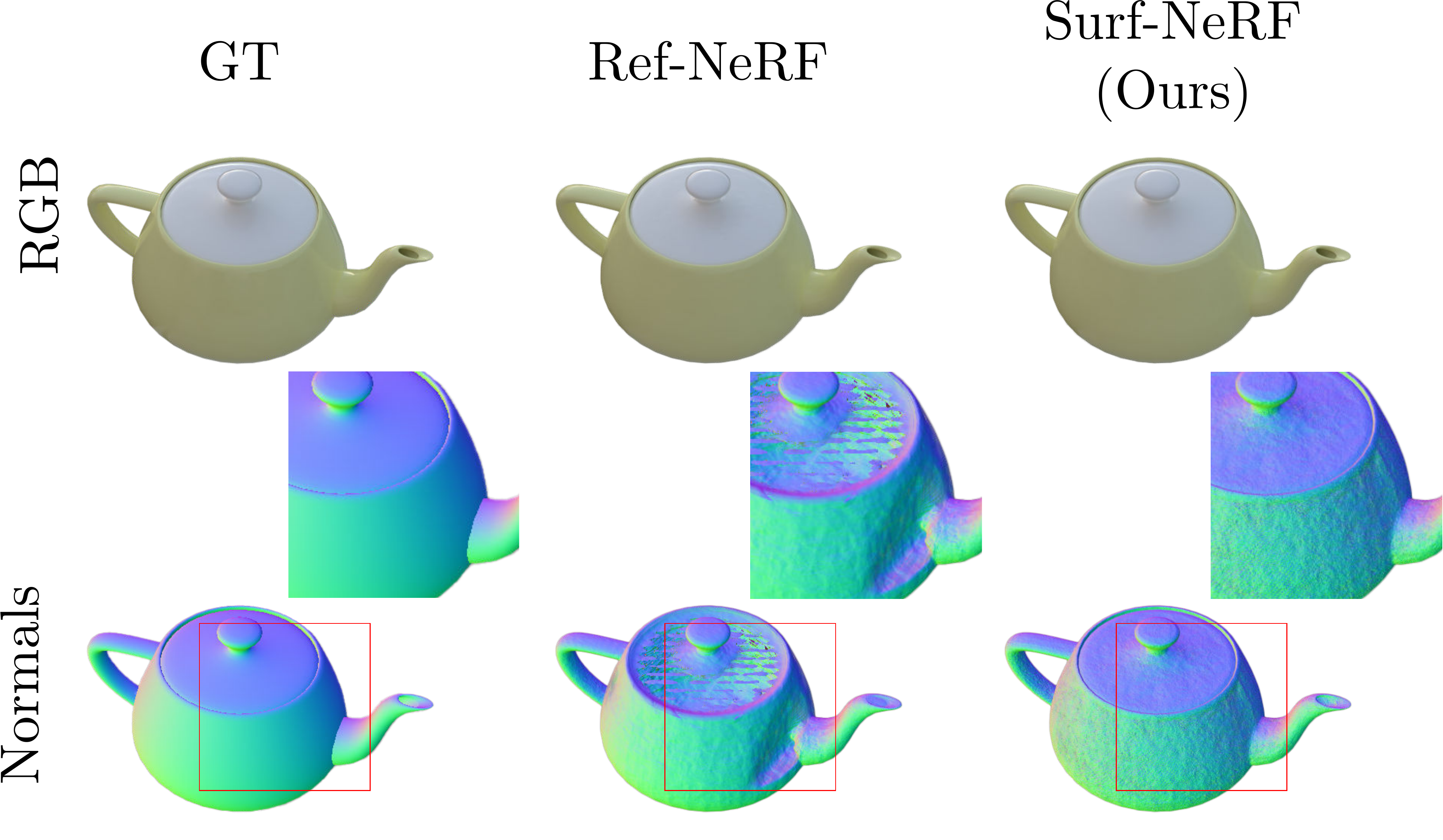}
    \caption{Geometry results on the non-grid based models. The surface regularisation can resolve errors in geometry which stem from shape-radiance ambiguity e.g. in regions with low-frequency textures on the teapot lid, without compromising visual fidelity. This attains substantially improved normals in these regions.}
    \label{fig:teapot}\vspace{-\floatsep}
\end{figure}

Figure~\ref{fig:teapot} depicts performance of positionally encoded models on specular surfaces with low texture. Surf-NeRF yields drastically better surface geometry and normals.

\begin{figure*}
    \centering
    \includegraphics[width=\linewidth]{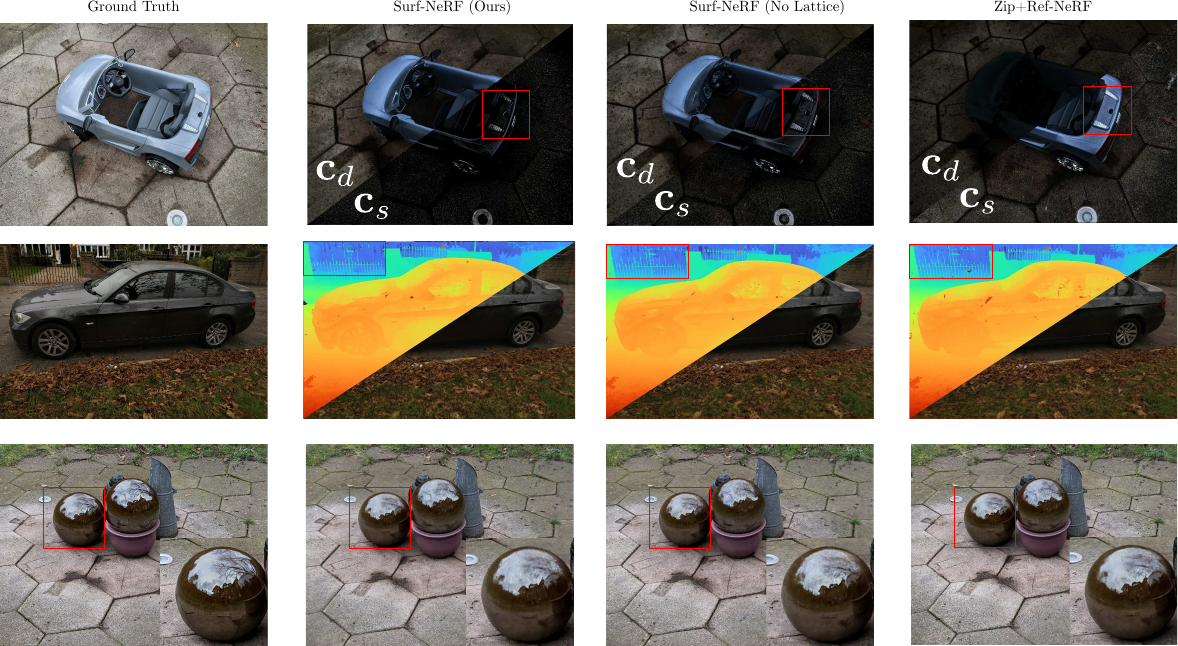}
    \caption{Results on the \textit{Shiny Real} dataset~\cite{verbin2022ref}. Surf-NeRF attains separation of diffuse and specular components $\vb{c}_d$ and $\vb{c}_s$ (top), a decreased reliance on floaters, more consistent depths, and qualitatively better representation of background content (middle), and comparable visual fidelity of reflective objects to other reflection parameterised NeRF variants.}
    \label{fig:shiny_real_supp}
\end{figure*}

Table~\ref{tab:shiny_real} demonstrates comparable visual fidelity of our proposed regularisation compared to the baseline approaches, with a small decrease overall owing to appearance and geometry being regularised in our curriculum learning framework. Figure~\ref{fig:shiny_real_supp} demonstrates qualitative performance on this dataset, illustrating improved separation, depth and comparable view fidelity.

\begin{figure*}
    \centering
    \includegraphics[width=\linewidth]{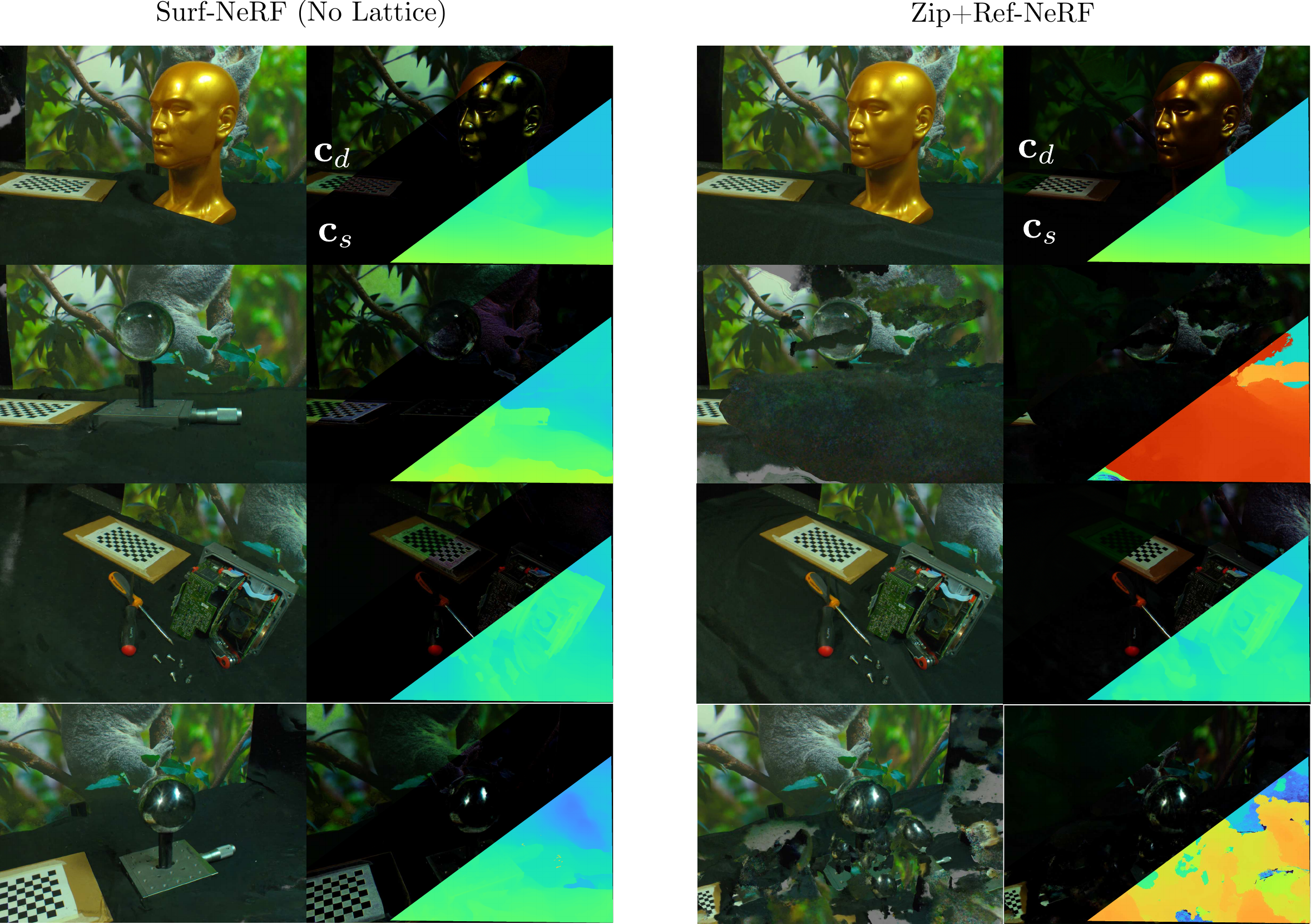}
    \caption{Our captured \textit{Koala} dataset in the no lattice case. We show improved separation of scene appearance between view-dependent and -independent content, more stable performance for scene content with high curvature and high view dependence. Renderings from models are on the left frame, with diffuse, specular and depth renderings on the right.}
    \label{fig:koala_all_scenes}
\end{figure*}

Table~\ref{tab:koala_supp} provides quantitative results for our captured dataset, showing aspects of improved visual fidelity in difficult scenes. These are illustrated accordingly in Figure~\ref{fig:koala_all_scenes}.

\begin{figure*}
    \centering
    \includegraphics[width=\linewidth]{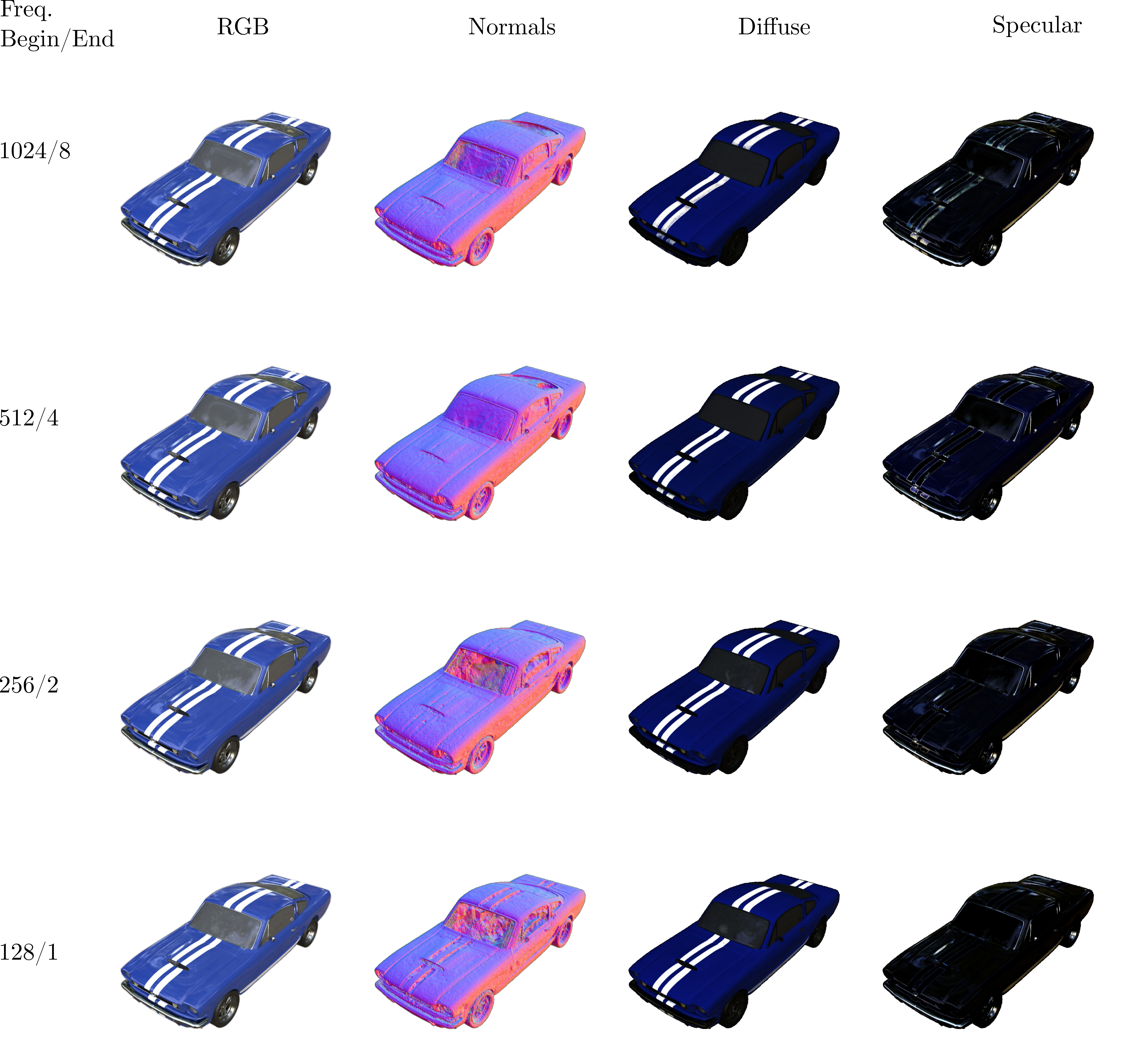}
    \caption{Visualisation of the curriculum learning parameter study in the Table~\ref{tab:currablation} on the car scene. As frequency increases, we see appearance separate more readily. We found that there were tradeoffs to the frequency of regularisation - as density is changed, the model must alter appearance to ``take up'' the changes to geometry. The 512/4 frequency was the pareto optimal state between strength of geometric improvement and the model's ability to alter appearance for the hash-based methods.}
    \label{fig:curriculum_ablation}
\end{figure*}

In Figure~\ref{fig:curriculum_ablation} we also show the impact of our curriculum ablation on the car scene, a comparatively difficult scene owing to low texture and high reflectively. We show that there is an optimal frequency to achieve improved normals and appearance separation, but higher frequencies more strongly separate this appearance at the detriment of geometry.

\begin{table}
\centering
\scriptsize
\caption{Per-Scene Results on the \textit{Shiny Objects}~\cite{verbin2022ref} dataset. Yellow, orange, red indicate third, second and first performing scores. Bold indicates best score for model type in disparity RMSE. We achieve improvements in all cases for disparity, overall improved normals and comparable visual fidelity results over the dataset.}

\begin{tabular}{clrrrr}
\hline
\multicolumn{1}{l}{} & Model            & PSNR & SSIM & MAE & RMSE\\
\hline
\multirow{7}{*}{Car}        & MipNeRF          & 26.99     & 0.921     & 47.44                         & 0.127\\
                            & MipNeRF+Diff     & 26.55     & 0.920      & 53.49                         & 0.130\\
                            & Ref-NeRF         & \cellcolor{red!50}31.17     & \cellcolor{red!50}0.957     & 15.49      & \textbf{0.123}\\
                            & Surf-NeRF (Reg. Only) & \cellcolor{orange!50}30.80      & \cellcolor{yellow!50}0.955     & \cellcolor{yellow!50}14.99 & \textbf{0.123}\\\cline{2-6}
                            & ZipNeRF          & 27.20     & 0.930      & 23.49                         & 0.207\\
                            & Zip+Ref-NeRF      & \cellcolor{yellow!50}30.37     & \cellcolor{yellow!50}0.955     & 16.37                         & 0.205\\
                            & Lattice+Ref-NeRF      & 30.30     & \cellcolor{orange!50}0.957     & 15.16                         & 0.209\\
                            & Surf-NeRF (No Lattice) & 29.82     & 0.953     & \cellcolor{orange!50}13.16                         & 0.198\\
                            & Surf-NeRF (Ours) & 29.85     & 0.955     & \cellcolor{red!50}11.08                         & \textbf{0.168}\\ \hline
\multirow{7}{*}{Coffee}     & MipNeRF          & 31.36     & 0.966     & 31.10                         & 0.124\\
                            & MipNeRF+Diff     & 31.45     & 0.966     & 38.97                         & 0.120\\
                            & Ref-NeRF         & \cellcolor{red!50}33.79     & \cellcolor{orange!50}0.973     & 13.50                         & \textbf{0.115}\\
                            & Surf-NeRF (Reg. Only) & \cellcolor{orange!50}33.56     & \cellcolor{yellow!50}0.972     & 12.54                         & \textbf{0.115}\\\cline{2-6}
                            & ZipNeRF          & 30.79     & \cellcolor{red!50}0.977     & 13.48                         & 0.207\\
                            & Zip+Ref-NeRF      & 32.56     & 0.971     & \cellcolor{yellow!50}11.95                         & 0.203\\
                            & Lattice+Ref-NeRF      & \cellcolor{yellow!50}32.80     & 0.972     & \cellcolor{red!50}9.93                         & 0.199\\
                            & Surf-NeRF (No Lattice) & 31.99     & 0.968     & 14.79                         & \textbf{0.185}\\ 
                            & Surf-NeRF (Ours) & 32.48     & 0.970     & \cellcolor{orange!50}11.57                         & 0.205\\ \hline
\multirow{7}{*}{Helmet}     & MipNeRF          & 29.03     & 0.943     & 75.28                         & 0.117\\
                            & MipNeRF+Diff     & 27.95     & 0.934     & 72.18                         & 0.112\\
                            & Ref-NeRF         & 29.31     & 0.952     & 40.96                         & \textbf{0.109}\\
                            & Surf-NeRF (Reg. Only) & 30.09     & 0.958     & 26.68                         & \textbf{0.109}\\\cline{2-6}
                            & ZipNeRF          & 27.07     & 0.945     & 24.96                         & 0.223\\
                            & Zip+Ref-NeRF      & \cellcolor{yellow!50}32.37     & \cellcolor{yellow!50}0.975     & 10.42    & 0.193\\
                            & Lattice+Ref-NeRF      & 32.31     & \cellcolor{orange!50}0.976     & \cellcolor{yellow!50}10.22  & 0.218\\
                            & Surf-NeRF (No Lattice) & \cellcolor{orange!50}32.93     & 0.974     & \cellcolor{orange!50}8.91                         & \textbf{0.192}\\ 
                            & Surf-NeRF (Ours) & \cellcolor{red!50}35.61     & \cellcolor{red!50}0.983     & \cellcolor{red!50}4.49                         & 0.200\\ 
                            \hline
\multirow{7}{*}{Teapot}     & MipNeRF          & 45.90      & 0.996     & 67.07                         & 0.152\\
                            & MipNeRF+Diff     & 45.62     & 0.996     & 69.10                         & 0.145\\
                            & Ref-NeRF         & \cellcolor{red!50}46.24     & \cellcolor{red!50}0.997     & 15.82                         & \textbf{0.136}\\
                            & Surf-NeRF (Reg. Only) & 45.76     & 0.996     & 16.07                         & \textbf{0.136}\\\cline{2-6}
                            & ZipNeRF          & \cellcolor{orange!50}46.11     & \cellcolor{red!50}0.997     & 9.10                          & 0.192\\
                            & Zip+Ref-NeRF      & \cellcolor{yellow!50}46.09     & \cellcolor{red!50}0.997     & 7.62                          & \textbf{0.157}\\
                            & Lattice+Ref-NeRF      & 46.08     & \cellcolor{red!50}0.997     & \cellcolor{yellow!50}6.52                         & 0.174\\
                            & Surf-NeRF (No Lattice) & 44.23    & 0.996     & \cellcolor{orange!50}6.17              & \textbf{0.157}\\
                            & Surf-NeRF (Ours) & 44.46    & 0.996     & \cellcolor{red!50}5.31              & 0.168\\\hline
\multirow{7}{*}{Toaster}    & MipNeRF          & 23.17     & 0.890      & 59.97                         & 0.104\\
                            & MipNeRF+Diff     & 22.63     & 0.864     & 66.27                         & 0.104\\
                            & Ref-NeRF         & \cellcolor{red!50}25.52     & \cellcolor{red!50}0.977     & 43.68                         & \textbf{0.102}\\
                            & Surf-NeRF (Reg. Only) & \cellcolor{orange!50}24.76     & 0.911     & 41.20                         & \textbf{0.102}\\\cline{2-6}
                            & ZipNeRF          & 23.91     & 0.911     & \cellcolor{yellow!50}26.33                         & 0.216\\
                            & Zip+Ref-NeRF      & \cellcolor{yellow!50}24.29     & \cellcolor{orange!50}0.921     & 27.16                       & 0.215\\
                            & Lattice+Ref-NeRF      & \cellcolor{yellow!50}24.17     & \cellcolor{orange!50}0.922     & 27.06                      & 0.209\\
                            & Surf-NeRF (No Lattice) & 21.67     & 0.879     & \cellcolor{orange!50}23.72                        & \textbf{0.198}\\
                            & Surf-NeRF (Ours) & 20.76     & 0.869     & \cellcolor{red!50}20.55                       & 0.202\\
                            \hline
% \multirow{4}{*}{\STAB{\rotatebox[origin=c]{90}{Pos.Enc.}}}    & MipNeRF          & \cellcolor{yellow!50} 31.29     & \cellcolor{yellow!50} 0.943     & \cellcolor{yellow!50} 56.17                         & 0.125                     \\
%                             & MipNeRF+Diff     & 30.84     & 0.936     & 60.00                         & \cellcolor{orange!50} 0.122                     \\
%                             & Ref-NeRF         & \cellcolor{red!50} 33.21     &  \cellcolor{red!50}0.971     & \cellcolor{orange!50} 25.89                         & \cellcolor{red!50} 0.117                     \\
%                             & Surf-NeRF (Ours) & \cellcolor{orange!50} 32.99     & \cellcolor{orange!50} 0.958     & \cellcolor{red!50} 22.30                         & \cellcolor{red!50} 0.117                     \\ \hline
%                         \multirow{3}{*}{\STAB{\rotatebox[origin=c]{90}{Hash}}}    & ZipNeRF          & \cellcolor{yellow!50} 31.02     &\cellcolor{yellow!50}  0.952     & \cellcolor{yellow!50} 19.47                         & \cellcolor{red!50} 0.209                     \\
%                             & Zip+Ref-NeRF      & \cellcolor{red!50} 32.76     & \cellcolor{red!50} 0.963     & \cellcolor{orange!50}15.68                         & \cellcolor{yellow!50} 0.232                     \\
%                             & Surf-NeRF (Ours) & \cellcolor{orange!50} 32.70     & \cellcolor{orange!50} 0.960     &  \cellcolor{red!50} 15.62                         & \cellcolor{orange!50} 0.211 \\
%                             \hline
\end{tabular}\label{tab:perscene_objects}
\end{table}

\begin{table}
\centering
\caption{\textit{Shiny Real}~\cite{verbin2022ref} dataset results. Yellow, orange, red indicate third, second and first performing scores. Applying our regularisation does not significantly degrade visual fidelity compared to unregularised variants.}
\scriptsize
\begin{tabular}{l l r r}
\hline
Scene & Model & PSNR & SSIM \\
\hline
\multirow{7}{*}{Garden Spheres} & MipNeRF & 22.37 & 0.452 \\
 & MipNeRF+Diff & \cellcolor{red!50}22.34 & 0.460 \\
 & Ref-NeRF & \cellcolor{yellow!50}22.17 & 0.413 \\
 & Surf-NeRF (Reg. Only) & \cellcolor{orange!50}22.33 & 0.443 \\ \cline{2-4}
 & ZipNeRF & 21.68 & \cellcolor{orange!50}0.539 \\
 & Zip+Ref-NeRF & 21.69 & \cellcolor{red!50}0.540 \\
 & Surf-NeRF (No Lattice) & 20.51  & 0.486 \\
 & Surf-NeRF (Ours) & 21.50  & \cellcolor{yellow!50}0.523 \\
 \hline
\multirow{7}{*}{Sedan} & MipNeRF & 24.71 & 0.600 \\
 & MipNeRF+Diff & 24.51 & 0.594 \\
 & Ref-NeRF & \cellcolor{orange!50}24.80 & 0.602 \\
 & Surf-NeRF (Reg. Only) & \cellcolor{yellow!50}24.72 & 0.614 \\\cline{2-4}
 & ZipNeRF & \cellcolor{red!50}25.99 & \cellcolor{red!50}0.730 \\
 & Zip+Ref-NeRF & 23.92  & \cellcolor{orange!50}0.687 \\
 & Surf-NeRF (No Lattice) & 23.71 & \cellcolor{yellow!50}0.684 \\
 & Surf-NeRF (Ours) & 24.26  & 0.676 \\
 \hline
\multirow{7}{*}{Toycar} & MipNeRF & \cellcolor{red!50}24.09 & 0.576 \\
 & MipNeRF+Diff & \cellcolor{yellow!50}24.00 & 0.576 \\
 & Ref-NeRF & 23.99 & 0.572 \\
 & Surf-NeRF (Reg. Only) & \cellcolor{orange!50}24.06 & 0.571 \\\cline{2-4}
 & ZipNeRF & 23.23 & \cellcolor{yellow!50}0.601\\
 & Zip+Ref-NeRF & 23.24 & \cellcolor{orange!50}0.609 \\
 & Surf-NeRF (No Lattice) &  23.20 & \cellcolor{red!50}0.620  \\
 & Surf-NeRF (Ours) &  23.36 & 0.600  \\
\hline
\end{tabular}~\label{tab:shiny_real_supp}
\end{table}

\begin{table}
\centering
\caption{Surf-NeRF results on our captured \textit{Koala} dataset. Bold indicates best performing scores. Regularisation helps the reflection-parameterised model around complex geometry and appearance on this dataset.}
\scriptsize
\begin{tabular}{l l r r}
\hline
Scene & Model & PSNR & SSIM \\
\hline
\multirow{3}{*}{Gold Head} & ZipNeRF & \textbf{29.00} & 0.680\\
 & Zip+Ref-NeRF & 28.34 & 0.681 \\
 & Surf-NeRF (No Lattice) & 27.62 & 0.680 \\
 & Surf-NeRF (Ours) & 26.61 & \textbf{0.683} \\
 \hline
\multirow{3}{*}{Shiny Ball} & ZipNeRF & \textbf{28.59} & 0.641 \\
 & Zip+Ref-NeRF & 15.47 & 0.435 \\
 & Surf-NeRF (No Lattice) & 28.17 & \textbf{0.646} \\
 & Surf-NeRF (Ours) & 27.78 & 0.643 \\
 \hline
\multirow{3}{*}{Lidar Guts} & ZipNeRF & 27.42 & 0.664 \\
 & Zip+Ref-NeRF & \textbf{27.55} & 0.665 \\
 & Surf-NeRF (No Lattice) & 26.84 & \textbf{0.666} \\
 & Surf-NeRF (Ours) & 25.88 & 0.642 \\
 \hline
\multirow{3}{*}{Crystal Capture} & ZipNeRF & 26.14 & \textbf{0.665} \\
 & Zip+Ref-NeRF & 23.40 & 0.577 \\
 & Surf-NeRF (No Lattice) & 26.33  & 0.627 \\
 & Surf-NeRF (Ours) & \textbf{27.17}  & 0.648 \\
\hline
\end{tabular}~\label{tab:koala_supp}
\end{table}

% \section{Videos}
% In addition to static results, we have attached to the supplement videos of renderings using our proposed approach, and those of the Zip+Ref-NeRF baseline method. These illustrate clearly the aspects of improved separation of view dependent and independent appearance, and surface normals and geometry. 
%\input{sec/4_results}
%\input{sec/5_conclusions}
% {
%     \small
%     \bibliographystyle{ieeenat_fullname}
%     \bibliography{main}
% }